%% file: main.tex
\documentclass[12pt]{article}

\input{1_header/packages.tex}

\begin{document}
\input{1_header/title.tex}
\restoregeometry

\section*{Abstract}
\input{1_header/abstract.tex}
\pagebreak

\section*{Acknowledgments}
\input{1_header/acknowledgments.tex}

\pagebreak

\setcounter{tocdepth}{5}
\tableofcontents
\pagebreak

\section{Introduction} \label{sect:intro}
\input{1_header/introduction.tex}

\section{Background} \label{sect:background}

\subsection{Reinforcement learning} \label{sect:rl}
\input{2_body/background/rl.tex}

\subsection{Deep reinforcement learning} \label{sect:deeprl}
\input{2_body/background/deeprl.tex}

\subsection{Generative adversarial networks} \label{sect:gans}
\input{2_body/background/gans.tex}

\section{Related work} \label{sect:relatedwork}

\subsection{Model-based reinforcement learning} \label{sect:model}
\input{2_body/relatedwork/imaginationrl.tex}

\subsection{GANs in reinforcement learning} \label{sect:gansrl}
\input{2_body/relatedwork/gansrl.tex}

\section{Imaginative framework for data efficiency} \label{sect:framework}
\input{2_body/framework.tex}

\section{Imagination module structure for GAIRL} \label{sect:imagination}
\input{2_body/imagination.tex}

\section{Experimental setup} \label{sect:setup}

\subsection{Environments}
\input{2_body/setup/environments.tex}

\subsection{Implementation} \label{sect:implementation}
\input{2_body/setup/implementation.tex}

\subsection{Evaluation metrics} \label{sect:metrics}
\input{2_body/setup/metrics.tex}

\section{Results} \label{sect:results}
\input{2_body/results.tex}

\section{Discussion} \label{sect:discussion}
\input{2_body/discussion.tex}

\section{Conclusion} \label{sect:conclusion}
\input{2_body/conclusion.tex}

\bibliographystyle{agsm}
\bibliography{bibliography}
\pagebreak

\begin{appendices}
\section{Numerical results for data efficiency} \label{app:results}
\input{3_appendices/results.tex}
\pagebreak

\section{Computational efficiency analysis} \label{app:compeff}
\input{3_appendices/computations.tex}
\pagebreak

\section{Generative hyperparameters for MNIST} \label{app:mnistgen}
\input{3_appendices/hyperparameters.tex}
\pagebreak

\end{appendices}
\end{document}

%% file: 1_header/packages.tex
\usepackage[utf8]{inputenc}
\usepackage[x11names,dvipsnames,svgnames,table]{xcolor}

\usepackage[export]{adjustbox}
\usepackage{afterpage}
\usepackage{graphicx}
\usepackage{subcaption}
\usepackage{algorithm}
\usepackage[noend]{algpseudocode}
\usepackage{lipsum}
\usepackage{parskip}
\usepackage[normalem]{ulem}
\usepackage{hyperref}       
\usepackage{url}            
\usepackage{booktabs}       
\usepackage{amsfonts}       
\usepackage{amsmath}    
\usepackage{color}
\usepackage{natbib}
\usepackage{tocbibind}
\usepackage{todonotes}
\usepackage{float}
\usepackage[nodayofweek,level]{datetime}
\usepackage[group-separator={,}]{siunitx}

\DeclareMathOperator*{\argmax}{argmax}
\DeclareMathOperator*{\EX}{\mathbb{E}}

\usepackage[titletoc,title,header]{appendix} 

\usepackage[a4paper]{geometry}
\usepackage{lastpage} 

%% file: 1_header/title.tex
\begin{titlepage}

\thispagestyle{empty}
\setlength\headheight{0pt} 
\begin{center}

\begin{center}
\includegraphics[width=0.25\linewidth]{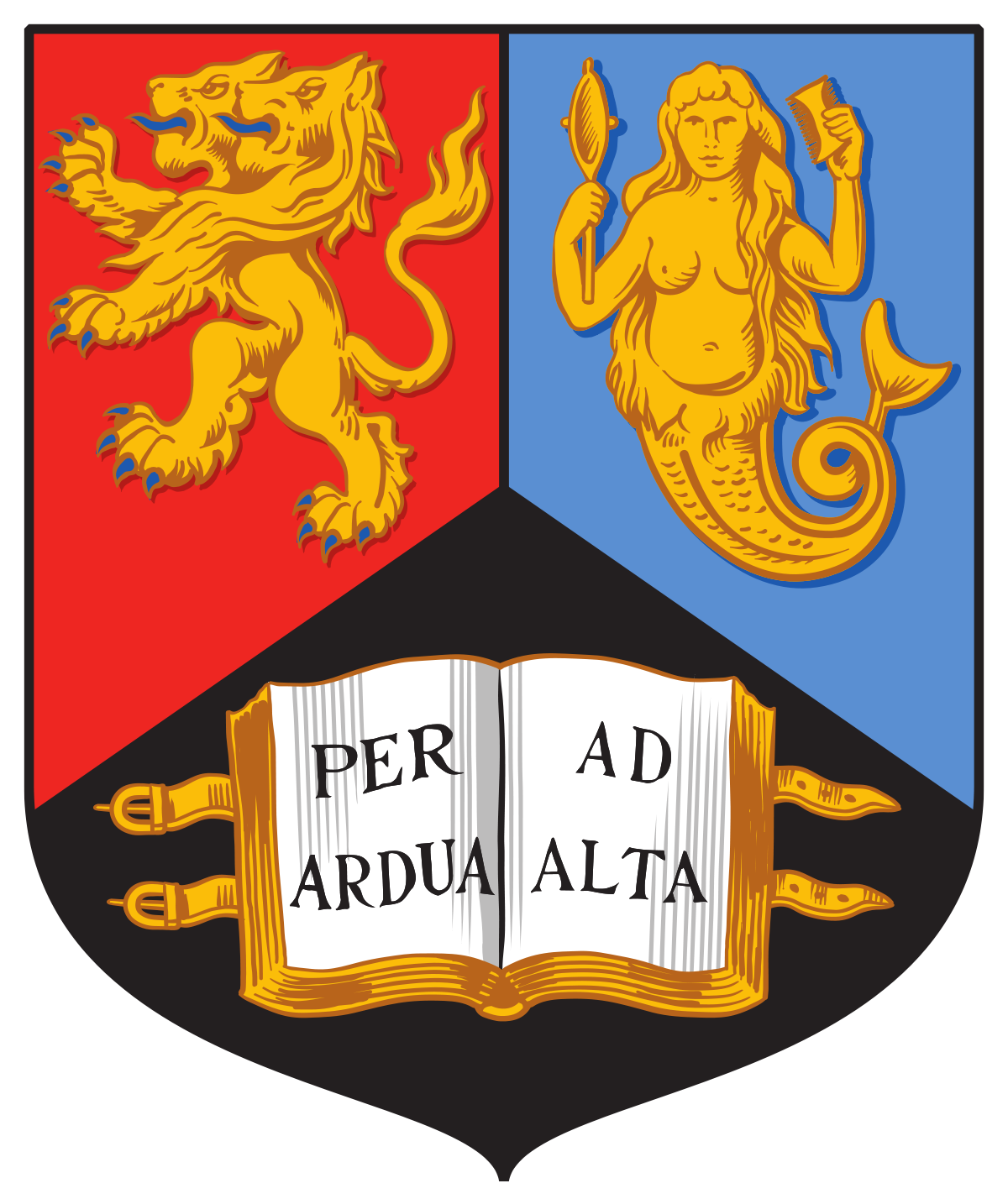}           
\end{center}	

        \vspace{0.25cm}
        {\scshape\LARGE University of Birmingham \par}
        \vspace{0.25cm}
        {\scshape\Large Undergraduate Thesis\par}
        \vspace{0.5cm}

        {\Large\bfseries Generative Adversarial Imagination for Sample Efficient Deep Reinforcement Learning\par}
        
        \vspace{0.5cm}
        {\Large\itshape Kacper P. Kielak\par}
        BSc Artificial Intelligence and Computer Science\\
        Student ID: 1698133
        \vspace{0.25cm}

\vspace{1cm}
Supervised by\par
Dr. Per Kristian Lehre \\
School of Computer Science, University of Birmingham\par
\vspace{1.5cm}
\large
{\monthname[4], 2019}

\end{center}

\clearpage
\restoregeometry
\end{titlepage}

%% file: 1_header/abstract.tex
Reinforcement learning has seen great advancements in the past five years. The successful introduction of deep learning in place of more traditional methods allowed reinforcement learning to scale to very complex domains achieving super-human performance in environments like the game of Go or numerous video games.

Despite great successes in multiple domains, these new methods suffer from their own issues that make them often inapplicable to the real world problems. Extreme lack of data efficiency, together with huge variance and difficulty in enforcing safety constraints, is one of the three most prominent issues in the field. Usually, millions of data points sampled from the environment are necessary for these algorithms to converge to acceptable policies.

This thesis proposes novel Generative Adversarial Imaginative Reinforcement Learning algorithm. It takes advantage of the recent introduction of highly effective generative adversarial models, and Markov property that underpins reinforcement learning setting, to model dynamics of the real environment within the internal imagination module. Rollouts from the imagination are then used to artificially simulate the real environment in a standard reinforcement learning process to avoid, often expensive and dangerous, trial and error in the real environment.

Experimental results show that the proposed algorithm more economically utilises experience from the real environment than the current state-of-the-art Rainbow DQN algorithm, and thus makes an important step towards sample efficient deep reinforcement learning.

%% file: 1_header/acknowledgments.tex
I would like to express my great gratitude to Dr Per Kristian Lehre for his invaluable guidance over the past year. His out-of-the-box suggestion during the planning phase of the project allowed me to put together different ideas that finally led to the work presented in this thesis.

In addition, I would like to thank Dr Lukasz Kaiser from Google Brain for finding the time to answer my questions regarding his recent findings.

%% file: 1_header/introduction.tex
One of the most prominent dilemmas in the field of artificial intelligence is to produce fully independent agents that learn optimal behaviour and develop over time purely by trial and error interaction with the surrounding environment. A mathematical framework that encapsulates the problem of these autonomous systems is reinforcement learning (RL) \citep{sutton1998rl}. Although over the past few years exceptional progress has been made in devising artificial agents that can learn and solve problems in a variety of domains using RL approaches \citep{arulkumaran2017survey}, these techniques are still not ideal. They require an immense amount of non-optimal interaction with the real environment before they begin to operate acceptably well and they do not efficiently adapt to new tasks, even within the identical environmental setting \citep{irpan2018rlfails}.

So far RL researchers were concentrating on mastering games like backgammon, chess, go, or various video games. In these settings dynamics of the environment are either entirely known and thus can be simulated (rules of the board games), or they can be queried and reset infinitely many times without any additional costs (video games). It allowed producing infinite amounts of data for the agent to learn. However, these kinds of conditions are rare in the real world. Dynamics of the environment are usually unknown and are too involved to approximate using rule-based methods. Often, we also cannot let the agent do millions of arbitrary trial-and-error live experiments freely. 

It is simple to imagine the use of reinforcement learning agent to optimise user experience on the website. Every bad decision in the real environment may result in a loss of an unsatisfied customer. Millions of such choices, before the agent converges to the optimal policy is too big of a risk for any company. Another example can be an autonomous car accustomed to driving in a specific country that suddenly finds itself in another country with a completely different driving culture. A human could quickly adapt to the new reality, but a current state-of-the-art reinforcement learning agent would require enormous amounts of experience first, highly increasing a probability of a severe accident.

Model-based reinforcement learning algorithms promise to solve this obstacle by using known dynamics of the environment to analyse probable scenarios. The agent can imagine various circumstances, and learn or reason based on them, without actually executing expensive trial-and-error exploration in the real world. Use of internal forecasts of the world for decision making and reasoning was deeply examined within the neuroscience community \citep{tolman1948cogmaps, hassabis2007hippimage, schacter2012memory}. It has been demonstrated to exist within the learning process of humans and several animals \citep{pfeiffer2013futpaths, leinweber2017vispred}. Again, it is manageable from the RL perspective when we fully understand the model of the environment as exhibited by the AlphaGo \citep{silver2017alphago}. However, as discussed earlier, we do not always know the exact specifics of the environment and, frequently, we have no prior knowledge regarding its dynamics at all.  

One plausible solution to that difficulty could be learning the model of the environment instead. Some work has been done already on the subject. Most prominently, \citet{oh2015ataripred} and \citet{leibfried2016rewardstatepred} showed that the dynamics of the environment can be modelled with very high accuracy. Nonetheless, although learned 'imaginative' model helped to improve outcomes in environments that require long-term planning, it did not significantly reduce the size of the system exposure required for training a well-performing agent \citep{racaniere2017i2a}. This brings about the following questions:
\begin{itemize}
    \item Can learning the imaginative model of the environment be more data efficient than learning an optimal policy?
    \item If so, can the learned imagination fulfil the promise of sample efficient model-based RL in settings where dynamics of the real environment are unknown?
\end{itemize}

This study subsequently answers both of the questions. It combines a few recent and a few less recent ideas from the field to do so. It hypothesises that recent advancement in the generative adversarial networks architecture \citep{goodfellow2014gan}, and inherent to the RL setting Markov property can provide a positive answer to the first question. Furthermore, it considers that the potential use of imagination within the structure similar to the Dyna-Q algorithm \citep{sutton1990dyna} may indeed profoundly improve sample efficiency of RL in unknown environments.

Following from the introduction, this thesis is structured as follows: Section \ref{sect:background} provides scientific background and basic theoretical fundamentals necessary for full understanding of the conducted research. Section \ref{sect:relatedwork} gives an overview of already existing studies that are relevant to this topic. Section \ref{sect:framework} introduces novel Generative Adversarial Imaginative Reinforcement Learning (GAIRL) algorithm developed to test the above-stated hypothesis. Section \ref{sect:imagination} then goes into details of methods that were used to efficiently create accurate imagination, i.e. learn the model of the environment. Section \ref{sect:setup} describes an experimental setting used to evaluate newly proposed GAIRL algorithm and compare it to the current state-of-the-art. Section \ref{sect:results} presents the qualitative results that are further discussed in section \ref{sect:discussion}. Finally, section \ref{sect:conclusion} summarises the work carried out.

%% file: 2_body/background/rl.tex
Reinforcement learning is the problem of learning an optimal policy (i.e. behaviour) for a given environment \citep{sutton1998rl}. RL is formalised by Markov decision processes (MDPs). An MDP can be formulated as a tuple $(S, A, T, R, \gamma)$ where $S$ is a (discrete or continuous) set of possible states, $A$ is a (discrete or continuous) set of allowed actions, $T$ is a transition probabilities function, $R$ is a (deterministic or stochastic) reward function, and $\gamma \in (0; 1)$ is a discount factor for future rewards (controls agent's time preference). At any given time $t$, the reinforcement learning agent observes an environment state $s_t \in S$ and selects an action $a_t \in A$. Then, the reward $r_{t+1} = R^{a_t}_{s_t}$ is returned from the environment, and the environment moves to the state $s_{t+1} \in S$ with transition probability $T^{a_t}_{{s_t}{s_{t+1}}} = P(s_{t+1}|s_t, a_t)$. $T$ and $R$ fully describe the dynamics of the environment. 

\begin{figure}[htb]
    \centering
    \includegraphics[width=0.75\textwidth]{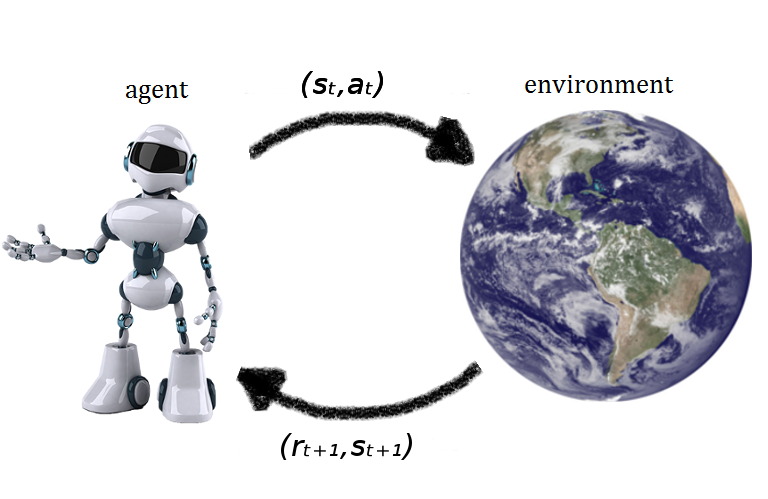}
    \caption{Reinforcement learning setting}
    \label{fig:rl}
\end{figure}

A critical characteristic of MDPs is that it follows the Markov assumption. Namely, at any point in time $t$, history of encountered states $s_0, s_1, \dots, s_{t-1}, s_t$ can be simplified to the last state $s_t$ without any loss in information, i.e:.
\begin{equation} \label{eq:mdp}
    P(s_{t+1} | s_t) = P(s_{t+1} | s_1,\dots,s_t)
\end{equation}
It is an assumption required by RL algorithms. Unfortunately, this property is not always observable in the real world. We can often circumvent its absence by clever preprocessing of the state space $S$ or by translating this lack of information to additional stochasticity on top of the transition and reward functions. However, even with these countermeasures, lack of Markov assumption may limit the range of optimal strategies that our agent can learn.

The agent interacts with an MDP by selecting actions accordingly to the policy $\pi$ that maps states to the probability distribution over A. The goal of obtaining an optimal policy can be formulated as learning a policy $\pi^*$ that maximises the state value function $V^\pi : S \to \mathbb{R}$ (Bellman equation):

\begin{equation}
    V^\pi(s) = \sum_{a \in A}~ \pi(a|s)(R^a_s+\gamma \sum_{s' \in S}~ T^a_{s{s'}} V^\pi(s')) \label{eq:stateval}
\end{equation}

From which we can also define state-action value function $Q^\pi : S \times A \to \mathbb{R}$:

\begin{equation}
    Q^\pi(s, a) = R^a_s+\gamma \sum_{s' \in S}~ T^a_{s{s'}} \sum_{a' \in A} \pi(a'|s') Q^\pi(s', a') \label{eq:stateactval}
\end{equation}

There is always at least one optimal policy and it satisfies/entails the following:
\begin{equation}
    \mathop{\forall}_{\pi} \mathop{\forall}_{s \in S} V^{\pi^*}(s) \geq V^\pi(s)
\end{equation}
\begin{equation}
    V^{\pi^*}(s) = \max_{a \in A} (R^a_s + \gamma \sum_{s' \in S}~ T^a_{s{s'}} V^{\pi^*}(s'))
\end{equation}
\begin{equation}
    Q^{\pi^*}(s, a) = R^a_s+\gamma \sum_{s' \in S}~ T^a_{s{s'}} \max_{a' \in A} Q^{\pi^*}(s', a')
\end{equation}

The setting could potentially be directly solved employing simple linear algebra: $V^{\pi^*} = (I - \gamma T)^{-1}R$; however, it works exclusively for finite-state MDPs and its time complexity is $O(n^{2.4})$ \citep{coppersmith1990matrix}. It is computationally infeasible for large and complex environments that are encountered in a vast majority of conditions. Therefore, it is rarely used, and most of the focus is given to algorithms that leverage sampling of the environment to approximate the optimal solution.

There are three different types of RL algorithms:
\begin{itemize}
    \item Value-based - explicitly learn state-action value function \eqref{eq:stateactval} of the environment and use it to infer optimal policy $\pi(s) = \argmax_{a \in A} Q(s, a)$. E.g. Q-learning \citep{watkins1992qlearning}.
    \item Policy-based - do not learn state value function explicitly but instead directly learn policy mapping $\pi : S \to A$. E.g. REINFORCE \citep{williams1992reinforce}.
    \item Actor-critic - combines learning policy (actor) and value function (critic). The actor makes choices about actions but it is updated by the feedback from the critic who then directly interacts with the environment. E.g. natural actor-critic \citep{peters2005natac}.
\end{itemize}
Value-based methods are usually the simplest and the fastest. The downside is that they do not work in continuous action spaces or when we want to learn stochastic policy.  Policy-based methods have better convergence qualities and can handle more complex policies and action spaces but are usually inefficient and converge to local minima. Actor-critic methods try to connect the advantages of both.

These traditional algorithms allowed for a big advancement in computer science creating artificial agents that have reached human-level performance in multiple games (e.g. backgammon  \citep{tesauro1995tdgammon}) without any prior knowledge. They work flawlessly for smaller state and action spaces where state value function \eqref{eq:stateval} and/or state-action value function \eqref{eq:stateactval} can be expressed by the look-up table. A tabular representation of these functions, however, is often not viable due to too big or continuous domains. 

%% file: 2_body/background/deeprl.tex
To allow RL to scale to more involved situations, one can propose using function approximators (such as linear regression, neural networks, or Bayesian methods) to approximate state-action value function $Q$ or policy $\pi$. Unfortunately, recursive updates of the functions \eqref{eq:stateval} \eqref{eq:stateactval} and lack of independent and identically distributed variables (future states/rewards highly depends on current state and action) in the RL setting break most of the assumptions expected by standard machine learning methods. Therefore, trials of plugging non-linear function approximators into the existing algorithms in the place of a tabular representation were resulting in complete divergence and failure.

Because it limited scalability of RL to many of the real-world problems, considerable amount of work has been done to solve this obstacle. Recently, \citet{mnih2015dqn} introduced Deep Q-Network (DQN), an RL algorithm that for the first time was capable of leveraging black-box properties of deep learning. They bypassed the dilemma of recursive updates and lack of \textit{i.i.d.} by proposing two adjustments to the standard Q-learning algorithm \citep{watkins1992qlearning}.

\textit{Replay buffer} - rather than instantly learning from sampled data that is highly correlated, algorithm stores $N$ most recent tuples $(s_t, a_t, r_{t+1}, s_{t+1})$ in a replay buffer $D$. When updating the value function $Q$, it uses a random mini-batch of samples from $D$ to estimate gradients. It reduces the correlation between samples by breaking their ordering.

\textit{Target network} - instead of updating $Q$ function directly with itself, the algorithm maintains two distinct networks: the online network $Q$ and the target network $\widehat{Q}$. $\widehat{Q}$ is simply a fixed snapshot of $Q$ taken every $C$ updates. The agent determines actions in a conventional manner accordingly to the network $Q$, but $Q$ is updated using a revised Bellman equation:
\begin{equation}
    Q(s, a) = R^a_s+\gamma \sum_{s' \in S}~ P^a_{s{s'}} \sum_{a' \in A} \pi(a'|s') \widehat{Q}(s', a')
\end{equation}
It stabilises the whole learning process and avoids exploding gradients by partially eliminating recursiveness in network updates.

DQN achieved super-human level performance on Atari 2600 games from the Arcade Learning Environment (ALE) \citep{bellemare2013atari} with raw pixels as input alone. This contribution opened an enormous amount of opportunities for RL. A lot of new algorithms and DQN improvements followed.

\citet{schaul2015prioritized} enhanced DQN slightly increasing its data efficiency by applying prioritised experience replay to more frequently replay more informative samples. \citet{vanhasselt2016ddqn} devised Double DQN that extends DQN to the double Q-learning method \citep{hasselt2010doubleq} that addresses an overestimation bias of the standard  Q-learning. \citet{wang2015duelling} introduced a new duelling network architecture specifically for the value-based RL that outperformed conventional supervised machine learning architectures. \citet{bellemare2017distributional} produced DQN-based algorithm that, alternatively to learning scalar state-action value function $Q$, learns a categorical distribution of the future returns $Z$ whose expectation is $Q$ proving both theoretically and experimentally that this procedure improves the original DQN algorithm. \citet{hessel2017rainbow} then consolidated all improvements enumerated in this paragraph (and a few smaller ones) into an algorithm called Rainbow DQN. It is the current RL state-of-the-art for discrete action spaces.

However, DQN, as a deep variant of Q-learning (value-based method), does not work in case of continuous actions space. Hence, it was soon merged with deterministic policy gradient methods \citep{silver2014dpg} to create the Deep Deterministic Policy Gradient (DDPG) actor-critic algorithm \citep{lillicrap2015ddpg} solving this issue. Recently, following the success of the Rainbow DQN, multiple incremental refinements suitable to the DDPG were consolidated forming Distributed Distributional Deep Deterministic Policy Gradient (D4PG) algorithm \citep{barth2018d4pg}. D4PG is the current state-of-the-art algorithm for complicated, continuous action space settings.

Although, the introduction of deep neural networks for the first time enabled RL algorithms to solve extremely complex problems and surpass humans at many levels, all of them suffer from tremendous data inefficiency. Rainbow DQN requires around 15 million frames of interaction with the real environment to match median human performance. It corresponds to over 8 days of constant play at the regular rate of 20 frames per second. It requires a total of full 200M frames to reach its peak performance (over three full months). In contrast, median human performance is defined as a score achieved by a person after merely 15 minutes of training beforehand. 

As explained in the introduction, this is not an enormous problem when training agent to master board or video games as computational power is relatively cheap nowadays. Three months of 20 frames per second can be shortened to 1 hour of $\sim\num{45000}$ frames per second given powerful enough infrastructure. However, it makes deep RL inapplicable to any other problem where obtaining samples of experience comes with potential additional costs like losing unsatisfied customers or causing an accident.

%% file: 2_body/background/gans.tex
Recently, \citet{goodfellow2014gan} introduced generative adversarial networks (GANs). They became a successful and widespread tool for data generation, profoundly over-performing previously used methods. Contrary to standard approaches that primarily focused on minimising L1 \eqref{eq:l1} or L2 \eqref{eq:l2} loss between a generated output and a real output on the individual level, GANs make it possible to work on the data distribution level minimising a difference between a generated data distribution and a real data distribution instead. 
\begin{equation}
    L1 = \sum_{i=1}^n |x_{generated} - x_{true}| \label{eq:l1}
\end{equation}
\begin{equation}
    L2 = \sum_{i=1}^n (x_{generated} - x_{true})^2 \label{eq:l2}
\end{equation}

GANs work by defining two separate networks. The generator $G : Z \to X$ that maps a noise vector $z \in Z$ coming from the noise distribution $p_z$ onto the data space $X$, and the discriminator $D : X \to [0, 1]$ that maps an input from the data space $X$ to the probability of the input coming from the real data distribution $p_{data}$. Both networks play a two-player minimax game with the objective:
\begin{equation}
    \min_G\max_D \EX_{x \sim p_{data}}[\log (D(x))] + \EX_{z \sim p_z}[\log (1 - D(G(z)))]
\end{equation}
As theoretically proven in the original study, this minimax game is equivalent to minimising the Jensen-Shanon (JS) divergence between $p_{data}$ and $p_g$:
\begin{equation}
    JS(p_{data} || p_g) \propto \EX_{x \sim p_{data}} [\log \frac{p_{data}(x)}{p_{data}(x) + p_g(x)}] + \EX_{x \sim p_g} [\log \frac{p_g(x)}{p_{data}(x) + p_g(x)}]
\end{equation}
and thus, modelling $p_g$ to as closely resemble $p_{data}$ as possible.

Unfortunately, GANs are also known for their lack of stability in training, often causing situations where one of the networks starts to completely overwhelm the other. This results in diminishing gradients for both of the networks. They also tend to ignore certain spectrums of the distribution (mode collapse problem). Therefore, numerous researchers tried to stabilise the original GAN algorithm. One of the most successful improvements was replacing JS divergence with the Earth-Mover distance (1st Wasserstein metric) creating the Wasserstein GAN \citep{arjovsky2017wgan}:
\begin{equation}
     \min_G\max_D \EX_{x \sim p_{data}}[D(x)] + \EX_{z \sim p_z}[-D(G(z))]
\end{equation}
Now, the discriminator (in the paper called critic) is no longer trying to predict if the data comes from the real or fake distribution. It is rather providing an actual real-valued distance (as measured by the Earth-Mover metric) between the data generated by the generator and the data coming from the real distribution. Hence, the goal is not longer to balance both networks but to make sure that the critic can converge to the real distance before letting the generator to improve. It has been mathematically proven in the study that the Wasserstein GAN always converges given that the critic is infinitely more powerful than the generator.

The issue with the Wasserstein GAN is that it constructs its minmax value function using the Kantorovich-Rubinstein duality \citep{cedric2008krduality}. Therefore, to be theoretically and practically sound, the critic needs to represent values coming from the set of 1-Lipschitz functions:
\begin{equation}
    f \text{ is 1-Lipschitz } \iff |f(x_1) - f(x_2)| \leq |x_1 - x_2| \label{eq:lipschitz}
\end{equation}
The original Wasserstein GAN enforced this limitation by clipping the weights of the critic network to space $[-c; c]$. Setting the $c$ hyperparameter is, however, a non-trivial task that introduced new instability problems.

Fortunately, \cite{arjovsky2017wgangrad} circumvented this concern by appending a gradient penalty to the final minimax objective instead of clipping weights of the critic:
\begin{equation}
    \min_G\max_D \EX_{x \sim p_{data}}[D(x)] + \EX_{z \sim p_z}[-D(G(z))] + \underline{\lambda \EX_{\widehat{x} \sim N(0, 1)}[(||\nabla_{\widehat{x}} D(\widehat{x})||_2 - 1)^2]}
\end{equation}
where $\lambda$ is a gradient penalty factor hyperparameter. $\lambda = 10$ in the original paper and $\lambda = 0$ recovers original Wassersetein GAN objective.

Unfortunately, the choice between the original GAN and the Wasserstein GAN with gradient penalty (WGANGP) is not that trivial. \citet{lucic2018ganscomp} showed that WGANGP is not necessarily outperforming standard GAN given suitable hyperparameters configuration. WGANGP also takes much longer to train because the critic, for every step of the generator, has to converge to the appropriate value of the Wasserstein distance fully. However, WGANGP's hyperparameters are much simpler to optimise. The choice between both usually comes down to the trade-off between computer power consumed by training and researcher/developer time spent on tuning the hyperparameters.

Furthermore, much research has been devoted to conditional GANs where the generator, rather than taking only a random noise $z$ as an input, takes another value $y$ based on which the final output is conditioned \citep{mirza2014cgan}. This enables us to not only generate arbitrary samples that follow the data distribution but also to have an influence on what precise spectrum of the distribution to obtain. This is particularly valuable when fitting GANs to the RL setting where the reward and next state tuple $(r_{t+1}, s_{t+1})$ is conditioned on the current state and chosen action pair $(s_t, a_t)$. The state-of-the-art in conditional GANs, especially in the area of computer vision, is PIX2PIX GAN that exhibited extraordinary results in multiple image-to-image translation problems \citep{isola2017p2pgan}.

\begin{figure}[htb]
    \centering
    \includegraphics[width=0.75\textwidth]{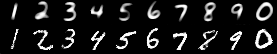}
    \caption{Comparison between GAN and L1-loss-based model on the classical MNIST dataset. Upper row presents results of an L1 loss generation, whereas lower row presents results of a GAN. The GAN produces very realistic images of handwritten data, indistinguishable from the original, due to successful data distribution approximation. L1 model, on the other hand, is only able to find images that minimise its mean error on the level of the individual, thus producing blurry results.}
    \label{fig:ganvsmlp}
\end{figure}

%% file: 2_body/relatedwork/imaginationrl.tex
As briefly discussed in section \ref{sect:intro}, one of the promises of the model-based reinforcement learning is to drastically improve the sample efficiency of reinforcement learning. The agent with access to the transition matrix $T$ and reward function $R$ (section \ref{sect:rl}) could internally reason about potential scenarios and outcomes without performing risky exploration in the real environment.  Unfortunately, the specific characteristics of the RL environment are usually unknown, and hence these methods cannot be directly applied.

\subsubsection{Learning the model - imagination} \label{sect:modellearning}

One plausible solution to that difficulty could be learning the model of the environment instead. The idea of learning the model of the environment from the sampled experience when its dynamics are not fully known is one of the most important concepts adapted in this thesis. However, this notion is not new. Current investigation in this area concentrates on application of variational autoencoders (VAEs) \citep{kingma2013vae}, recurrent neural networks (RNNs) \citep{medsker1999rnn}, and/or Bayesian methods. \citet{lenz2015deepmpc} introduced a novel robot control algorithm leveraging RNN that predicted the position of robot's parts. \citet{bellemare2013bayes} factors the state space to decompose model learning into smaller, more manageable sub-problems and applies Bayesian inference to predict future states. 

A substantial breakthrough in learning the model of the environment in the RL setting and the algorithm that is most widely applicable was proposed by \citet{oh2015ataripred}. The paper presented two novel Encoding-Transformation-Decoding architectures to learn the transition probabilities function $T$. They first encode high-dimensional state $s_t$ using convolutional network \citep{lecun1999cnn}, then the transformation conditioned on $a_t$ is performed to convert a high-level encoding of the of the current state to a high-level encoding of the next state, and finally decoding using deconvolutional network \citep{zeiler2010decnn} is executed to decode high-level next-state features into the full representation of the next state $s_{t+1}$. The first architecture employs a simple feed-forward method and takes a fixed history of states as an input. The other takes advantage of LSTM cells \citep{hochreiter1997lstm} to capture the most relevant features from the past. 

This work was later extended through a straightforward modification to support modelling of reward function $R$, and thus, to able to learn the full model of the environment \citep{leibfried2016rewardstatepred}. It was also incrementally improved by \citet{chiappa2017recatari} by the alteration of the original architecture and exploration of a few novel ideas.

Nevertheless, these model generation techniques have two major areas for improvement: 
\begin{itemize}
    \item They do not utilise Markov property \eqref{eq:mdp} fully, treating the model learning as a highly sequential and non-stationary problem instead of transforming it into much easier, stationary scenario. It makes capturing all intrinsics of the environment much harder to learn. 
    \item They utilise L1 or L2 objective to train the model and predict future states. L1/L2 loss penalises each difference between prediction and ground truth uniformly. Thus, it struggles to prioritise small but significant features over large but meaningless (e.g. slight ball location change over reduced saturation of a black background in Pong Atari game) and often produces blurry results. This is a well-known problem when applying standard classification/regression deep learning models for generative tasks. 
\end{itemize}

\subsubsection{Applications of imagination} \label{sect:imagapp}

Historically, there has been a significant amount of work devoted to model-based reinforcement learning methods that took advantage of known environment dynamics. Firstly, as already mentioned in section \ref{sect:rl}, having that knowledge allows to directly solve the RL setting using simple linear algebra. If state/action is too intricate for that, we can apply traditional planning techniques. Additionally, Monte Carlo tree search \citep{browne2012mcts} can be deployed for more reliable evaluation of the future rewards. Several model-based techniques like DQN-cognate methods coupled with prior knowledge about the intrinsic model of the environment have already yielded outstanding effects \citep{silver2016alphago, silver2017alphago}.

Sadly, using these techniques directly on the imagination (approximated real environment) has been shown to perform very poorly \citep{talvitie2014error1, talvitie2015error2}. It is because approximations, by their nature, introduce a certain level of error. This error quickly compounds over time when performing long environment rollouts into the future. Utilising this approximation to estimate policy using planning-like methods is directly translating this compounded error into estimated policy causing it to be, in the best case, sub-optimal.

One of the first algorithms that managed to do that successfully was Dyna-Q \citep{sutton1990dyna}. It used a look-up table for imagination modelling. The imagination was later used to simulate and replace the real environment. Unfortunately, it did not scale to more complex problems due to the use of a finite look-up table. The high-level structure of the algorithm proposed in this thesis is inspired by the traditional Dyna-Q and can be partially viewed as its expansion to bigger and/or continuous state spaces.

Most recently, \citet{racaniere2017i2a} successfully employed model learning techniques described in section \ref{sect:modellearning} by combining them with the standard model-free methods in the decision making process. Their results showed improvements in the effective exploration and in the environments where long-term planning is crucial, even, when the imagination of the agent was far from perfect. Nevertheless, it did not significantly reduce the size of the system exposure necessary for training a well-performing agent.

Recent work that most closely resembles this thesis was done by \citet{venkatraman2016efficient1} and \citet{gu2016efficient2}. They also try to improve sample efficiency of RL by using Dyna-Q inspired algorithms. However, both of the studies adopt simple, often linear, methods to train the imagination. Hence, they can only be applied to specific domains where such methods are successful. Additionally, their data efficiency is not specifically better than the current model-free state-of-the-art Rainbow DQN \citep{hessel2017rainbow}.

%% file: 2_body/relatedwork/gansrl.tex
GANs and RL are usually treated as separate fields of machine learning research by the community. There was not much attention to combining them or employing advances from one domain to improve the other. Recent work by \citet{pfau2016ganrl} presented a deep connection between GANs and actor-critic RL trying to encourage both communities to learn from each other. 

A few studies to utilise GAN architecture to devise novel RL algorithms followed. \citet{doan2018ganqlearning} introduced GAN Q-learning, a model-free distributional alternative to the DQN algorithm, by using generative adversarial architecture to express Bellman update \eqref{eq:stateactval} implicitly. \cite{ho2016gail} obtained significant performance gains in imitation learning \citep{schaal2003imitation} by applying a novel Generative Adversarial Imitation Learning algorithm. \citet{henderson2018optgan} proposed an innovative OptionGAN architecture for inverse reinforcement learning \citep{ng2000inverse} improving results in one-shot transfer.

This thesis proposes the use of GANs to learn the dynamics of the environment and form the agent's imagination. Similar approach was presented by \citet{xiaoganmcts} and \citet{azizzadenesheli2018ganmcts}. They both examined learning the model of the environment for Atari games to, similarly to the AlphaGo \citep{silver2017alphago}, apply a combination of DQN with the Monte Carlo tree search to effectively discover the optimal policy. Both studies reported negative results due to the too short MCTS rollouts as proven in the former. \cite{azizzadenesheli2018ganmcts}, however, was able to remarkably efficiently learn a very accurate representation of model dynamics using generative adversarial approach.

%% file: 2_body/framework.tex
Building upon previous research, this thesis introduces Generative Adversarial Imaginative Reinforcement Learning (GAIRL) algorithm to answer questions posed in section \ref{sect:intro} and to improve overall data efficiency of deep RL. As briefly mentioned in section \ref{sect:relatedwork}, study by \citet{sutton1990dyna}  inspired the main structure of the novel algorithm. This structure is a focus of this section, without going much into details of the generative adversarial imagination. Transforming the imaginative framework specifically into the GAIRL algorithm is a topic of the next section. 

The imaginative core underpinning GAIRL is divided into two separate modules: the model-free module (MFM) and the imagination module (IM). It also makes use of the concept of memory $M$. The memory is simply an array storing previous real-environment experience, similarly to the replay buffer in the DQN-cognate algorithms that was described in section \ref{sect:deeprl}.

These modules are utilised across three distinct training phases. First is the model-free phase (MFP) that only makes use of the MFM, followed by the imagination training phase (ITP) where solely the IM is operated, finalised by the imagination-based phase (IBP) that combines both of the modules.

\subsection{Model-free module}

The MFM consists of a standard model-free reinforcement learning algorithm; in this study, it is the state-of-the-art Rainbow DQN. It is the core decision-making module that models policy mapping $\pi : S \to A$ of the agent. The advantage of the imaginative framework is that it can leverage any other existing model-free algorithm if it is more suited for a given task (e.g. D4PG in environments with continuous action space) or if a model-free algorithm that overperforms Rainbow DQN is introduced in the future.

\begin{figure}[htb]
    \centering
    \includegraphics[width=0.75\textwidth]{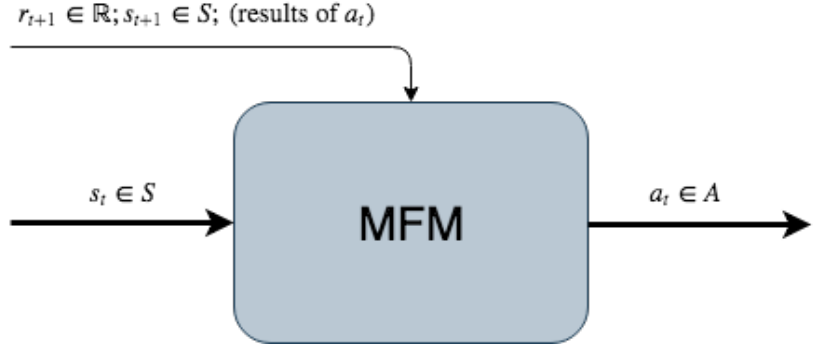}
    \caption{Model-free module. It is in the essence a standard RL algorithm working exactly as described in sections \ref{sect:rl} and \ref{sect:rl}. It represents policy $\pi(s_t) = a_t$ that is incrementally updated based on the consequences of performed actions ($r_{t+1}, s_{t+1}$).}
    \label{fig:mfm}
\end{figure}

\subsection{Imagination module}

The IM is a crucial part of the whole framework. It is a trainable system that can simulate the behaviour of the environment dynamics, i.e. accurately approximate the transition probabilities function $T$ and the reward function $R$, similarly to the approach presented by \citet{leibfried2016rewardstatepred}. Similarly to the MFM, any generative model can be employed in place of the IM. Nonetheless, this thesis focuses on a specially designed IM to optimise its data efficiency. In depth design of the IM to create the GAIRL algorithm can be found in section \ref{sect:imagination}. 

\begin{figure}[htb]
    \centering
    \includegraphics[width=0.75\textwidth]{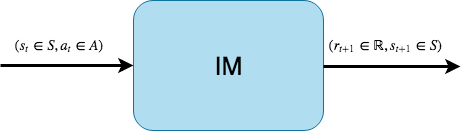}
    \caption{Imagination module. Its goal is to replace the real environment by modelling its dynamics (functions $T$ and $R$). Thus, its input/output exactly resembles input/output of the environment shown in \figurename{~\ref{fig:rl}}.}
    \label{fig:im}
\end{figure}

\subsection{Model-free phase}

The MFP mostly follows the standard RL procedure taking advantage of the model-free RL algorithm employed within the MFM. Similarly to the standard RL, it is based entirely on the real environment. The only difference is that, in addition to the use of experience samples returned by the environment to improve the policy $\pi$ encapsulated in the MFM, it also stores these samples within the agent's memory $M$. 

MFP is the only one out of three phases that requires the real environment to sample experience. Therefore, to limit the amount of the real experience that is required, it is also recommended to keep it as short as possible.

\begin{figure}[H]
    \centering
    \includegraphics[width=0.85\textwidth]{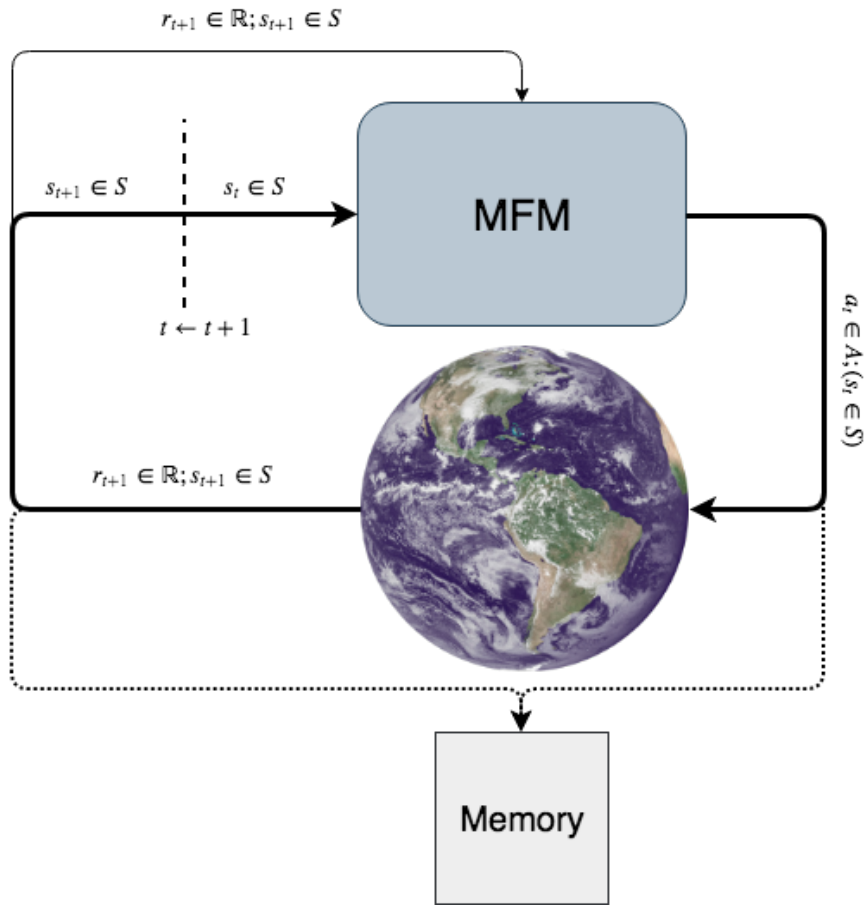}
    \caption{Model-free phase. Excluding the memory that stores transition tuples ($s_t, a_t, r_{t+1}, s_{t+1}$), the whole concept was already visualised in \figurename{~\ref{fig:rl}}. MFM given a state $s_t$ produces an action $a_t$. The pair ($s_t, a_t$) is fed to the real environment (technically $s_t$ is not produced by the agent nor fed into the environment, environment simply \textbf{is} in the state $s_t$). Environment moves into the next state $s_{t+1}$ and produces the reward $r_{t+1}$. The pair ($r_{t+1}, s_{t+1}$) is then given as a feedback to the agent and together with the pair ($s_t, a_t$) it is saved in the memory. Finally, the next state $s_{t+1}$ becames the new current state ($s_t$) and the whole process starts from the beginning.}
    \label{fig:mfp}
\end{figure}

\subsection{Imagination training phase}

Imagination training phase is focused on using transitions samples ($s_t, a_t, r_{t+1}, s_{t+1}$) stored in the memory $M$ to train the IM to accurately follow the dynamics of the real environment (approximate functions $T$ and $R$) in a purely supervised learning manner.

Length of the ITP does not affect the real environment at all as it exploits past, memorised, experience instead. Hence, it can, and it should to produce more optimal results, run until the IM fully converges to the representation of the data that is stored in the memory.

\subsection{Imagination based phase}
Having fully trained imagination, we can start leveraging it. The IBP is focused on improving the agent's policy by letting MFM train and rollout experimental scenarios on the IM instead of making potentially expensive trial and error in the real world. In fact, the MFM is not even aware of the fact that its learning process has been moved onto the 'fake' environment instead of the real one.

\begin{figure}[htb]
    \centering
    \includegraphics[width=0.85\textwidth]{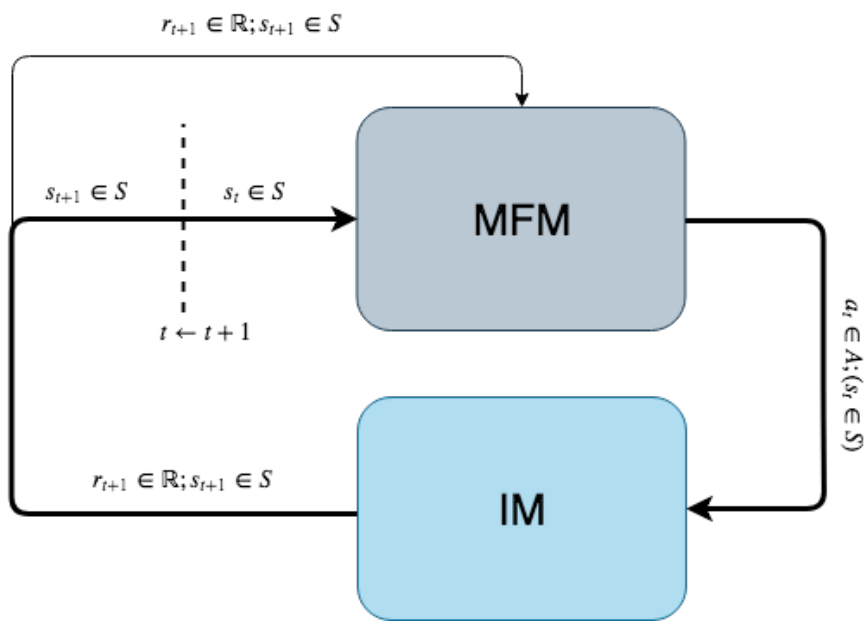}
    \caption{Imagination based phase. The concept works exactly like the one visualised in \figurename{ \ref{fig:rl}} or in \figurename{ \ref{fig:mfp}} (excluding memory). The only difference is that the real environment was replaced by the imagination module.}
    \label{fig:ibp}
\end{figure}

IBP, similarly to the ITP, should last until the agent's policy fully converges to the imaginative environment to extract as much signal, from experience gathered so far, as possible. Nevertheless, in practice, experiments focused on performing only three times more steps in the imagination than in the real environment, even, when the policy did not always fully converged during the imagination based phase.

\subsection{Summary}
The whole premise of the framework is to extract meaningful signal from the obtained experience as efficiently as possible. Imagination module serves as a way to generalise possible distribution of the world to create a safe, artificial, imaginative environment where the agent can learn and experiment without any risks associated with the actions in the real world.

The three phases work in a loop following Algorithm \ref{alg:gairl}. The loop part is crucial. In the first MFP, the agent usually performs completely random and experimental actions and is very likely to not reach more advanced environment states such as the next level in a video game. Going through the ITP and the IBP can allow it to master the first level of the game. However, it needs to go back to the real environment to explore newly reachable states to be able to imagine them in the second iteration accurately. This process should continue until the agent's policy fully converges to the given environment.

Convenient characteristic of the whole framework is that it can work for any model-free RL algorithm in place of the MFM and any generative module in place of the IM. Nevertheless, only the framework that specifically leverages GANs and Markov property for the data efficient IM is defined as GAIRL as explained in the next section.

\begin{algorithm}
\caption{Imagination for sample efficient reinforcement learning} \label{alg:gairl}
\begin{algorithmic}[1]
\Procedure{GAIRL}{$MFM, IM, env$}
    \State Initialise $MFM$
    \State Initialise $IM$
    \State Create and initialise $M$
    \While{true}
        \State Train $MFM$ on $env$ while collecting experience samples in $M$ (MFP)
        \If{$MFM$ converged on $env$}\\
~~~~~~~~~~~~~~\Return $MFM$ agent
        \EndIf
        \State Train $IM$ on the data from $M$ (ITP)
        \State Train $MFM$ on $IM$ (IBP)
    \EndWhile
\EndProcedure
\end{algorithmic}
\end{algorithm}

%% file: 2_body/imagination.tex
The framework described in section \ref{sect:framework} is based on the assumption that the IM can learn the dynamics of the real environment accurately, and efficiently enough. As discussed in section \ref{sect:modellearning}, there already exist some studies on this topic. Most prominently \citet{oh2015ataripred} used variational autoencoders in combination with recurrent neural networks to create a model that can predict the next state in Atari games conditioned on the current state and the chosen action. Its results were extremely accurate; however, it did not focus on sample efficiency.

Contrary to the previous studies, the IM of the GAIRL algorithm takes into account Markov property simplifying the whole setting to the straightforward supervised learning problem with an entirely stationary mapping $S \times A \to \mathbb{R} \times S$. It is, therefore, theoretically, much more data efficient, especially compared to the standard model-free policy learning that tries to learn a behaviour maximising expected cumulative reward, i.e. the extremely non-stationary sum of all future rewards. 

Although any supervised learning model that can fit the above-mentioned description could be used in the framework, certain algorithms are superior to others. As described in section \ref{sect:gans}, standard deep learning models based on the L1 or L2 loss do not perform well in generative tasks. In theory, the best performing architecture to approximate different high-dimensional generative distributions are generative adversarial networks \citep{goodfellow2014gan}. GANs should add another advantage over popular variational autoencoder approach. As mentioned in section \ref{sect:gansrl}, use of the PIX2PIX GAN \citep{isola2017p2pgan} for this setting already yielded remarkably good and sample efficient results in the very recent study \citep{azizzadenesheli2018ganmcts}.

Another essential characteristic of GANs is that they take random noise as an input (in addition to the conditional input in case of standard generative models). This quality provides an additional advantage over previously used methods in highly stochastic environments. It should allow GAN, if needed, to produce a set of diverse stochastic outputs for the same conditional input, instead of just a deterministic mean of possible outputs. 

\begin{figure}[H]
    \centering
    \includegraphics[width=0.99\textwidth]{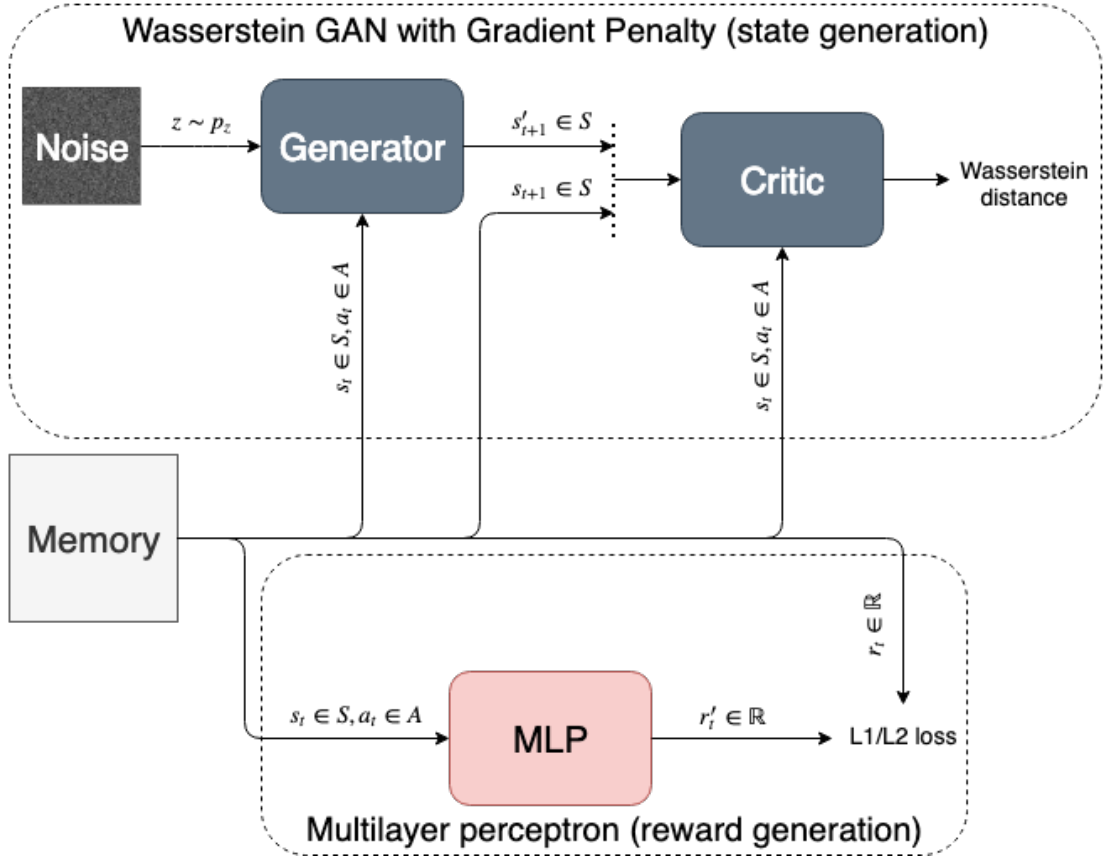}
    \caption{GAIRL's imagination training phase. For the state prediction, a tuple $(s_t, a_t)$ sampled from the memory serves as a conditional input for both generator and critic. The generator, given random noise $z$ and a conditional input, tries to generate a 'fake' next state $s_{t+1}'$ to as closely resemble the real next state $s_{t+1}$ as possible. Then, the critic, knowing conditional inputs for each of the possible fake and real states, calculates a Wasserstein distance between the generator conditional distribution $p_g$ and the real environment conditional distribution $p_{env}$ that the generator aims to reduce. For the reward prediction, a tuple $(s_t, a_t)$ is the standard input to the MLP. The MLP tries then to predict the following reward. The predicted reward $r_{t+1}'$ is then combined with the real reward $r_{t+1}$ to calculate loss that the MLP tries to minimise.}
    \label{fig:imganitp}
\end{figure}

\begin{figure}[htb]
    \centering
    \includegraphics[width=0.99\textwidth]{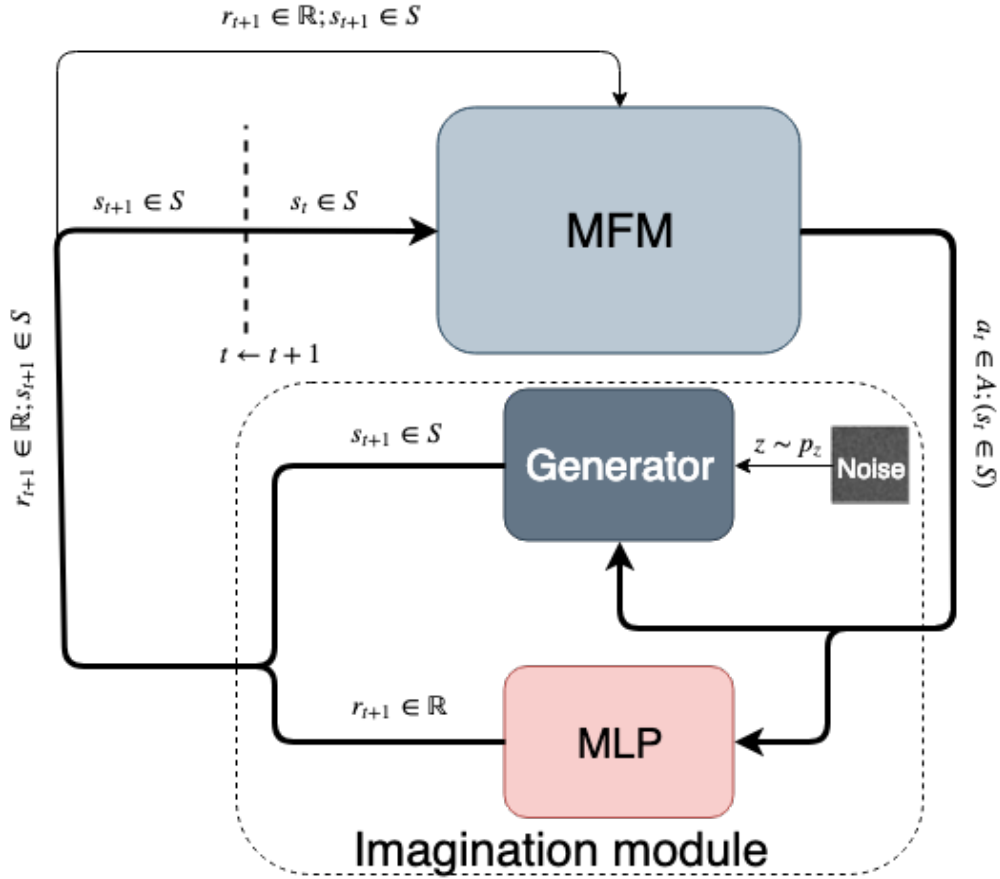}
    \caption{The state-predicting generator and the reward-predicting MLP plugged into the IBP. Both generator and MLP work as the imagination module.}
    \label{fig:imganibp}
\end{figure}

Although single GAN seems like a perfect match for the IM, in practice, it is easier to construct the IM using two separate deep learning models. One predicting the next state (modelling transition probabilities function $T$), another predicting the expected reward (modelling reward function $R$). In the proposed framework, the next state generation is handled by a GAN because state space is often highly dimensional structure perfectly suited for that architecture. Reward, however, is always represented by a single, real-valued, scalar. Therefore, it can be simplified to a simple regression problem. Hence, GAIRL employs a traditional L1/L2 loss based regression model for the reward generation task. The full internal structure of the IM module during the ITP can be seen in \figurename{ \ref{fig:imganitp}}. \figurename{ \ref{fig:imganibp}} shows how trained imagination is then used in the IBP.
 
The final crucial step towards a successful use of the IM in the GAIRL setting was deciding on starting states for the IM rollouts. Real environment simply starts in a particular state. IM, on the other hand, is just a combination of supervised learning models, it does not have any inherently associated internal state. States are merely inputs to the machine learning system. Therefore, in the IBP, whenever there is a need to start a new episode, a random state is sampled from the memory. Just then, having these ground truth initial state, the IM is used to simulate possible scenarios into the future in the standard manner.

%% file: 2_body/setup/environments.tex
To asses the capabilities of the algorithms, OpenAI Gym \citep{openaigym} was employed. It provides traditional and most popular reinforcement learning environments in an easy to use manner. For the past four years, the most important reinforcement learning benchmarks were Atari games from the Atari Learning Environment \citep{bellemare2013atari} (also provided by the OpenAI gym). Unfortunately, RL algorithms require very extensive hardware to converge to the optimal policies for these games (up to 100GB of RAM per single run of the algorithm). 

Therefore, due to high time constraints and limited resources, this study employs slightly simpler benchmarks from the classic control family of the problems. It mainly focuses on the MountainCar \citep{moore1990mountaincar}, and Acrobot \citep{sutton1996acrobot} environments. Although classic control environments are simpler and operate on lower dimensions, they were the standard benchmark in the reinforcement learning community for decades before Atari games were adopted. Nevertheless, in addition to showing the proof of concept of the GAIRL framework on the MountainCar and Acrobot, the plan is to continue the work to analyse the framework on currently adopted test environments.

\subsubsection{MountainCar}
\begin{figure}[htb]
    \centering
    \includegraphics[width=0.75\textwidth]{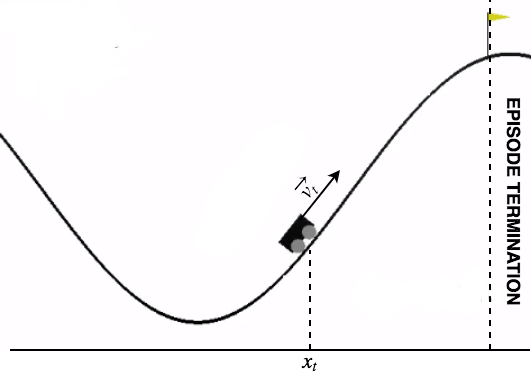}
    \caption{MountainCar environment.}
    \label{fig:mountaincar}
\end{figure}
MountainCar is based on a simple, low-dimensional setting. The agent controls a car that must drive up the steep slope. The car is not able to climb the hill directly. The agent has to learn to leverage the fact that it is situated in a valley and can use potential energy of the opposite slopes.

The state space is represented by two continuous variables: horizontal position of the car $x_t$, and velocity of the car across the car's axis $v_t$ ($s_t = (x_t, v_t) \in \mathbb{R}^2$). Action space is defined by a single choice: drive left, or drive right ($a_t \in \{-1, 1\}$). The agent gets a reward $r = -1$ for every time step until the episode is terminated (car drives up the right hill). An optimal policy should minimise the time it takes to reach the top of the hill, and thus maximise the cumulative reward obtained during the episode.

To solve the MountainCar environment, the algorithm has to be capable of handling continuous state space. Additionally, what makes it harder to learn is a very sparse reward signal. The agent has no information about the goal until it is reached. A meaningful signal occurs once per from 200 (close-to-optimal policy) to over 5000 (random policy) actions.

\subsubsection{Acrobot}
\begin{figure}[htb]
    \centering
    \includegraphics[width=0.75\textwidth]{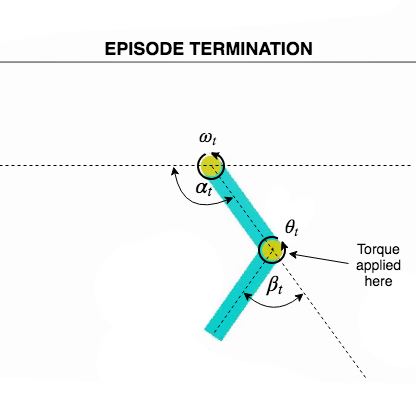}
    \caption{Acrobot environment.}
    \label{fig:acrobot}
\end{figure}

Acrobot, although still fairly low-dimensional, is harder and much more difficult task. The agent controls a two-link, under-actuated robot arm. The first (upper) joint, attached to the background, is out of control of the agent. The only controllable part is the torque of the lower joint of the robot. The goal is to balance the whole arm, so the tip of the second link swings above the episode termination line. 

Six continuous variables represent the state space: sinus and cosine of the angle $\alpha_t$ between the first link and a horizontal line; sinus and cosine of the angle $\beta_t$ between the first and the second link; an angular velocity $\omega_t$ of the first joint; and an angular velocity $\theta_t$ of the second joint. 
$$s_t = (\sin{\alpha_t}, \cos{\alpha_t}, \sin{\beta_t}, \cos{\beta_t}, \omega_t, \theta_t) \in [-1;1]^4 \times R^2$$
Action space is again defined by a single choice, although slightly modified: apply positive torque, no torque, or negative torque to the second joint ($a_t \in \{-1, 0, 1\}$). Similarly to the MountainCar, the agent gets a reward $r = -1$ for every time step until the episode finishes (tip of the second link swings above the termination line) to encourage policies that take the least amount of time to complete the episode.

%% file: 2_body/setup/implementation.tex
All algorithms were implemented using the Tensorflow machine learning library \citep{abadi2016tensorflow} with Python as a front-end programming language. Results and performance of the algorithms were captured and visualised using TensorBoard, a visualisation tool that is a part of the Tensorflow framework.

\subsubsection{Model-free reinforcement learning} \label{sect:rlimplement}
Implementation started with the standard Deep Q-Network code following the original paper \citep{mnih2015dqn}. The initial structure consisted of 2 hidden layers with 24 nodes each to accustom for simpler environments. Although the most popular activation function for hidden layers of deep neural networks is rectifier linear unit (ReLU) ($f(x) = \max(0, x)$) introduced by \citet{nair2010relu}, from my experience it fairly often causes 'dead neurons' problem, i.e. input to the ReLU is negative (thus output is a constant $0$) causing lack of gradient flowing through the node during the backpropagation. Therefore, from the beginning, DQN was employing leaky rectifier linear units ($f(x) = \max(\alpha x, x)$) with default parameter $\alpha_{lrelu} = 0.2$ \citep{maas2013lrelu}. Starting configuration was using random minibatches of 32 experience samples from the replay buffer and Adam optimisation technique \citep{kingma2014adam} to derive an optimal set of network weights.

It was then debugged and hypertuned until it was able to solve both, previously mentioned, environments easily. Most interesting changes performed during the hypertuning regarded the batch size and the optimisation technique. Initially, the agent was not able to even reach close-to-optimal policies. A single change from Adam optimiser to the standard stochastic gradient descent without any additional acceleration (SGD) was enough to fix the issue. Deep reinforcement learning is known to suffer from high instability. Momentum-based optimisation may often exacerbate this problem. Additionally, SGD has been recently shown to have better generalisation properties \citep{wilson2017sgdgood}, what may be another reason for its superiority in this setting. Following this change, the agent became able to reach optimal-like behaviour; however, it was quickly forgetting what it has learned and collapsing back to sub-optimal policies. The solution was to increase the batch size, from 32 to 256 strongly. A batch made out of only 32 samples was often not diverse enough to represent the actual signal coming from the environment, thus again intensifying instabilities of deep RL.

Afterwards, the state-of-the-art Rainbow DQN was written inheriting the DQN structure and following improvements described by \citet{hessel2017rainbow}. It followed the same debugging and hypertuning procedure. The final Rainbow DQN hyperparameters are presented in \tablename{ \ref{table:rainbow}}. The same Rainbow DQN was then used both as a baseline and within the model-free module of the GAIRL framework.

\begin{table}[ht]
\centering
    \begin{tabular}{|l|l|} \hline
    Hyperparameter                               & Value              \\ \hline
    Hidden layers                                & $[24, 24]$         \\
    Hidden layers activation                     & Leaky ReLU         \\
    Leakiness parameter ($\alpha_{lrelu}$)       & $0.2$              \\
    Dropout probability                          & $0$                \\
    Final layer activation                       & Linear             \\
    Optimiser                                    & Gradient descent   \\
    Learning rate ($\alpha_{lr}$)                & $5 \times 10^{-3}$ \\
    Gradient clipping                            & $1$                \\
    Discount factor ($\gamma$)                   & $0.99$             \\
    Exploration strategy                         & $\epsilon\text{-greedy}$ \\
    Exploration $\epsilon$ decay                 & $1 \to 0.05 \text{ (linear)}$ \\
    Exploration decay start                      & $\num{1000} \text{ steps}$ \\
    Exploration decay length                     & $\num{10000} \text{ steps}$ \\
    Replay buffer size                           & $\num{10000}$      \\
    Replay batch size                            & $256$              \\
    Prioritisation $\epsilon$                    & $1 \times 10^{-5}$ \\ 
    Prioritisation $\alpha$                      & $0.6$              \\ 
    Prioritisation $\beta$ decay                 & $0.4 \to 1 \text{ (linear)}$ \\ 
    Prioritisation decay length                  & $\num{50000}$      \\
    Noisy networks $\delta_0$                    & $0.5$              \\
    Multi-step returns $n$                       & $3$                \\
    Online network update frequency              & $4$                \\
    Target network update frequency              & $500$              \\
    \hline
    \end{tabular}
\caption{Final parameters of the Rainbow DQN agent.} 
\label{table:rainbow}
\end{table}

\subsubsection{Generative models} \label{sect:gansimpl}
Following the successful implementation, debugging, and hypertuning reinforcement learning agents, implementation of the second group of necessary algorithms~--~generative models~--~has started. All of the neural networks for the generative models were also initially configured to use Leaky ReLU activation for the hidden layers and Adam optimisation for training. 

In the first step, I implemented standard GAN based on the original work by \citet{goodfellow2014gan}. Similarly to the DQN, it was then debugged and hypertuned; however, this time using standard MNIST dataset \citep{lecun1999cnn}. At first, both generator and discriminator seemed like they entirely lack gradient to progress, yet none of the well known GAN issues occurred. Interestingly, the problem did not lay purely in standard hyperparameters of the networks but in the weights initialisation. Originally, GAN's weights were initialised according to the same distribution as weights of RL algorithms. In DQN, weights have to be initialised to values with a mean $\mu = 0$ and a very low standard distribution to avoid divergence and exploding gradients that can be caused by recursive updates. For GANs, however, these values were too low to produce any meaningful signal.  Increasing the standard deviation of initial weights distribution solved the difficulty. Unfortunately, networks imbalance problem that was described in detail in section \ref{sect:gans} followed. Discriminator started to completely overwhelm generator causing the gradient of both networks to vanish. The solution to that was decreasing the number of discriminator training steps for each generator training step from $k = 5$ to $k = 1$.

After the original GAN was implemented, both Wasserstein GAN (WGAN) and Wasserstein GAN with gradient penalty (WGANGP) were built on top. Because tuning weight clipping constant $c$ in WGAN can be extremely hard and mundane, after making sure there are no bugs in the code, focus moved to the WGANGP implementation, without fully hypertuning the standard WGAN. Sadly, the WGANGP did not perfectly work on the dataset with the same hyperparameters as the standard GAN. The reason behind it was that, unlike the discriminator in the standard GAN, the critic in the WGANGP has to be much more powerful than the generator so it can fully converge to the real value of the Wasserstein distance between the real and the generated data. Slightly increasing the network size of the critic and the number of critic steps per single generator step from $k = 1$ to $k = 10$ caused the algorithm to perform much better. Nevertheless, it was still suffering from high variance and problematic convergence. As hypothesised and proved by the next runs of the model, this was caused by too high learning rate in both of the networks.

Furthermore, after implementation of the GAN, WGAN, and the WGANGP based on the original papers, they were expanded to allow for a conditional generation as proposed by \citet{mirza2014cgan}. Conditional versions of these algorithms did not need any additional hypertuning to learn probabilistic distribution underlying the MNIST dataset adequately.

As described in section \ref{sect:imagination}, an L1/L2 loss based model is also necessary for the expected reward generation. Therefore, in addition to GANs, a standard multilayer perceptron (MLP), optimising for the mean absolute error (L1 loss), has been implemented. L1 objective, instead of the most popular mean squared error (L2), has been chosen because it had been shown to work better for generative tasks \citep{zhao2017l1better}. The MLP was also initially debugged and hypertuned to fit the MNIST dataset. The comparison between implemented WGANGP and MLP on the MNIST dataset can be seen in figure \figurename{ \ref{fig:ganvsmlp}}. It is a great example of the superiority of GANs over traditional models in high-dimensional generative tasks. 

What is also crucial, the property that all generative models followed equally, both for the MNIST dataset and as a part of the imaginative framework, was normalisation. Namely, all values from the data, both inputs and outputs, were scaled to fit the range $[0; 1]$. It further optimised their performance but also made it easier to interpret results. The fact that the output space is a range $[0; 1]$ is leveraged by employed evaluation metrics as described in section \ref{sect:metrics}.

Final hyperparameters of models that were optimised for the MNIST can be found in \appendixname{ \ref{app:mnistgen}}. Nonetheless, they are not important for the final results described in section \ref{sect:results}.

\subsubsection{GAIRL}
Once generative algorithms and the model-free RL agent had been implemented, putting together the GAIRL agent begun. As explained in sections \ref{sect:framework} and \ref{sect:imagination}, the model-free module consists of the RL agent, and two generative models take the place of the imagination. 

In place of the MFM, the final version of Rainbow DQN described in section \ref{sect:rlimplement} is used. Precisely the same network structure and hyperparameters were employed for two reasons: Firstly, to entirely make sure that experimental results show actual merit of the GAIRL method, and are not merely caused by varying hyperparameters between the algorithms. Secondly, to test the promise of GAIRL to work as a universal sample efficiency enhancement in a very modular, plug and play manner, that does not require any special tuning of the RL algorithms employed within the MFM.

Although one of the central premises of the research presented in this thesis is the advantage of GANs for the next state generation part of the imagination module, it was plausible that on chosen, less dimensional benchmarks they may actually perform worse than the standard L1/L2 loss approaches due to their higher level of complexity. Therefore, two different versions of the IM were implemented: an MLP-based imagination and a WGANGP-based imagination. The MLP-based version uses an MLP model for both next state and reward generation. Whereas WGANGP-based sticks to the standard GAIRL premise of using a GAN for the next state and an MLP for the reward. Also, this decision rendered a good comparison between Generative Adversarial Imaginative Reinforcement Learning and potentially simpler version of the framework. For both versions, the output of the MLP used for the reward generation is rounded to the closest integer (to accustom for discrete rewards in the experimental environments).

\begin{table}[h]
\centering
    \begin{tabular}{|l|l|} \hline
    Hyperparameter                               & Value              \\ \hline
    Hidden layers                                & $[512, 512]$       \\
    Hidden layers activation                     & Leaky ReLU         \\
    Leakiness parameter ($\alpha_{lrelu}$)       & $0.2$              \\
    Dropout probability                          & $0$                \\
    Final layer activation [generator]           & Tanh               \\
    Final layer activation [critic]              & Linear             \\
    Penalty coefficient ($\lambda$)              & 10                 \\
    Critic steps per one generator step ($k$)    & 10                 \\
    Optimiser                                    & Adam               \\
    Learning rate ($\alpha_{lr}$)                & $2 \times 10^{-4}$ \\
    First Adam decay rate ($\beta_1$)            & $0.5$              \\
    Second Adam decay rate ($\beta_2$)           & $0.9$              \\ \hline
    \end{tabular}
\caption{Final parameters of the state-generating WGANGP for the WGANGP-based imagination.} 
\label{table:gairlgan}
\end{table}

\begin{table}[h]
\centering
\hspace*{-0.1\linewidth}
\begin{subtable}[b]{.55\linewidth}
    \centering
    \begin{tabular}{|l|l|}
    \hline
    Hyperparameter                               & Value              \\ \hline
    Hidden layers                                & [512, 512]         \\
    Hidden layers activation                     & Leaky ReLU         \\
    Leakiness parameter ($\alpha_{lrelu}$)       & $0.2$              \\
    Dropout probability                          & $0$                \\
    Final layer activation                       & Linear             \\
    Optimiser                                    & Adam               \\
    Learning rate ($\alpha_{lr}$)                & $2 \times 10^{-4}$ \\
    First Adam decay rate ($\beta_1$)            & $0.9$              \\
    Second Adam decay rate ($\beta_2$)           & $0.999$            \\ \hline
    \end{tabular} 
    \caption{MLP used for the state generation in the MLP-based imagination.}
\end{subtable}%
\hspace*{0.05\linewidth} \begin{subtable}[b]{.55\linewidth}
    \centering
    \begin{tabular}{|l|l|}
    \hline
    Hyperparameter                               & Value              \\ \hline
    Hidden layers                                & [64, 64]           \\
    Hidden layers activation                     & Leaky ReLU         \\
    Leakiness parameter ($\alpha_{lrelu}$)       & $0.2$              \\
    Dropout probability                          & $0.25$             \\
    Final layer activation                       & Linear             \\
    Optimiser                                    & Adam               \\
    Learning rate ($\alpha_{lr}$)                & $2 \times 10^{-4}$ \\
    First Adam decay rate ($\beta_1$)            & $0.9$              \\
    Second Adam decay rate ($\beta_2$)           & $0.999$            \\ \hline
    \end{tabular} 
    \caption{MLP used for the reward generation in both MLP-based and WGANGP-based imagination.}
\end{subtable}
\captionsetup{width=1.15\linewidth}
\caption{Final parameters of multilayer perceptrons used for the imaginative framework.} 
\label{table:gairlmlp}
\end{table}

As described in section \ref{sect:gans}, an original GAN could be more computationally efficient than the WGANGP. However, a hard deadline on the thesis moved the focus on minimising the time necessary to tune hyperparameters, instead of minimising the computational efficiency. 

What is more, both environments presented in this study have deterministic dynamics, and so generative models do not require any inherent stochasticity. Therefore the random noise input to the WGANGP was omitted. All of the hyperparameters, both for WGANGP and MLPs, had to also be customised to the new setting. Tables \ref{table:gairlgan} and \ref{table:gairlmlp} show final hyperparameter choices for all of the generative models used for the final evaluation.

In practice, the imaginative framework also maintains two separate memory modules: one representing a training set, the other a test set. It is vital to analyse the capabilities of the imagination module thoroughly. The memory used for experiments consisted of $\num{200000}$ most recent samples from the environment, 80\% of which was used for training and 20\% was separated purely for assessing the IM. A critical characteristic of the memory was that it artificially increased the number of experience samples that result in terminal states by oversampling \citep{ling1998oversampling}. It was necessary due to a massive imbalance of terminal/non-terminal states in almost every RL setting. To do so, GAIRL keeps track of the mean length of episodes $\mu_{ep}$. Once the episode finishes (a transition sample with a terminal state occurs), the obtained sample is replicated $\lceil\mu_{ep}\rfloor$ many times.

As per algorithm \ref{alg:gairl}, three training phases, MFP, ITP, and IBP continue in a loop starting from the model-free phase. MFP was initially set to operate for $\rho_1 = \num{20000}$ steps in the real environment, ITP for $\rho_2 = \num{20000}$ stochastic gradient descent imagination training steps, and IBP for $\rho_3 = \num{60000}$ steps in the imaginative environment. However, initial experiments showed that longer ITP with $\rho_2 = \num{40000}$ steps better generalises on both environments and so the hyperparameter has been changed for the final version of the algorithm.

%% file: 2_body/setup/metrics.tex
Different metrics have been used to test different aspects of the algorithms. Precision \eqref{eq:prec} and recall \eqref{eq:recall} is employed to assess the performance of the reward generation (normalised reward in the chosen environment can be either $0$ or $1$). Precision describes a ratio of true positives (correctly generated $1$s) within all generated positives (all generated $1$s, no matter if correctly or not). Recall defines how many $1$s were correctly generated out of all true $1$s. They provide much more information about the real performance of the machine learning model than a standard accuracy metric, especially in situations of high-class imbalance that takes place in chosen environments (reward $0$ is extremely more common than $1$).
\begin{equation} \label{eq:prec}
    \text{precision} = \frac{\text{true positives}}{\text{true positives} + \text{false positives}}
\end{equation}
\begin{equation} \label{eq:recall}
    \text{recall} = \frac{\text{true positives}}{\text{true positives} + \text{false negatives}}
\end{equation}

State, on the other hand, follows fully continuous dynamics. Therefore, mean absolute error (MAE) \eqref{eq:mae} is used for state-generating models. MAE is simply an averaged L1 loss that calculates a mean absolute difference between a generated state and a ground truth state. It has been chosen over more popular mean squared error for the same reasons L1 loss has been chosen over L2 loss (see section \ref{sect:gansimpl}). Given that all outputs are normalised between $0$ and $1$, the value of $1 - \text{MAE}$ can also be referred to as accuracy. 
\begin{equation} \label{eq:mae}
    \text{MAE} = \frac{1}{n} \sum_{i=1}^n |x_{generated} - x_{true}|
\end{equation}

Finally, to evaluate data efficiency of algorithms (the main focus of this thesis), simply a mean reward from the $100$ most recent episodes, in regards to the number of steps performed in the real environment, is used.

%% file: 2_body/results.tex
First, imagination capabilities to accurately approximate real environment dynamics are presented. Then, focus moves onto data efficiency of different variations of the imaginative framework and the state-of-the-art Rainbow DQN. Furthermore, computational efficiency analysis of the proposed framework can be found in \appendixname{ \ref{app:compeff}}. All experiments were performed using algorithms and parameters described in section \ref{sect:implementation}, the same for both environments. Each of the showcased results is based on 15 independent runs of the algorithm. In all of the plots, lines represent a mean value for the runs. Opaque areas represent a standard deviation around the mean.

\subsection{Imagination performance} \label{sect:imagresults}

\subsubsection{Reward generation}

\begin{figure}[htb]
\centering
\begin{subfigure}[b]{.5\linewidth}
    \centering
    \includegraphics[width=1 \textwidth]{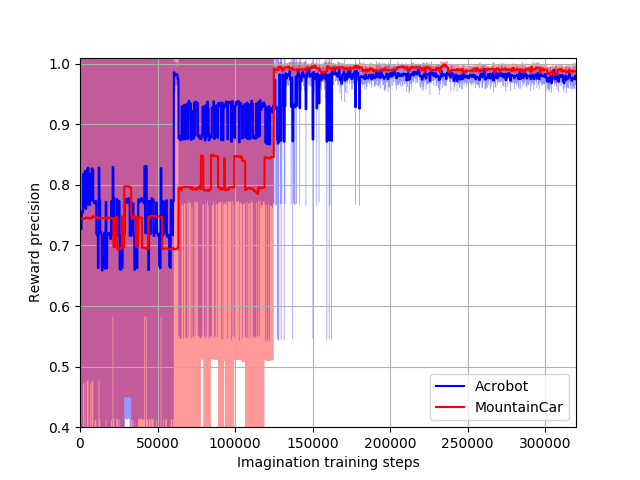}
    \caption{Recall on the test memory.}
\end{subfigure}%
\begin{subfigure}[b]{.5\linewidth}
    \centering
    \includegraphics[width=1 \textwidth]{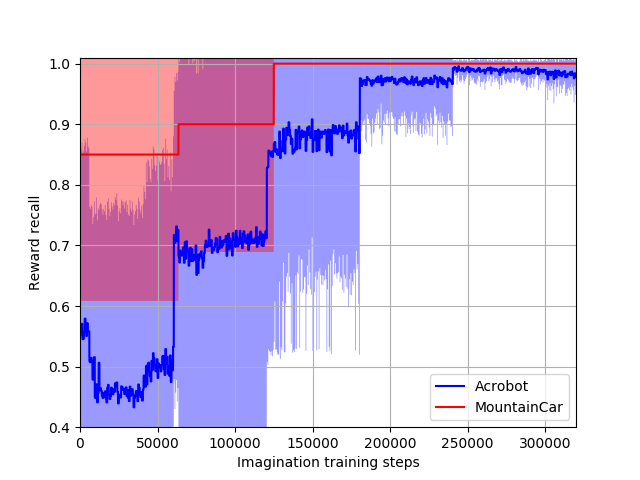}
    \caption{Precision on the test memory.}
\end{subfigure}
\caption{Performance of the reward-generating MLP imagination submodule on experimental environments. The x-axis represents training steps performed solely in the ITP (in other phases the performance stays constant). Y-axis shows the value of an appropriate metric.}
\label{fig:reward}
\end{figure}

\figurename{ \ref{fig:reward}} shows the performance of the reward imagination submodule in terms of precision and recall in both environments. Both metrics, besides recall of Acrobot reward that requires twice as much experience, converge after $\num{120000}$ steps. For MountainCar, as it is a simple environment, the MLP can easily reach over $99\%$ recall and precision in a few GAIRL iterations. Acrobot is a bit more challenging and even converged imagination's precision can drop below $99\%$. Although results of this magnitude are sufficiently accurate, they could potentially be improved if more time was spent on hypertuning the machine learning model.

What can be intriguing, are big jumps about every $\num{60000}$ steps in the performance of the network as the training progresses. They are caused by the the main GAIRL algorithm loop. For every GAIRL iteration, agent gathers $\num{20000}$ samples of experience from the real environment ($\rho_1 = \num{20000})$ to then train the imagination for $\num{60000}$ gradient descent steps ($\rho_2 = \num{60000}$). Therefore, each multiple of $\num{60000}$ marks a point after which more real data enters the process helping the model to better generalise to unseen samples. I.e. in the first $\num{60000}$ steps imagination learns only based on $\num{16000}$ samples ($80\%$ of $\num{20000}$ because another $20\%$ belongs to the test memory), in the $\num{60000} - \num{120000}$ period based on $\num{32000}$, in the $\num{12000} - \num{180000}$ based on $\num{48000}$, and so on.

\subsubsection{State generation}

\begin{figure}[H]
\centering
\begin{subfigure}[b]{.5\linewidth}
    \centering
    \includegraphics[width=1 \textwidth]{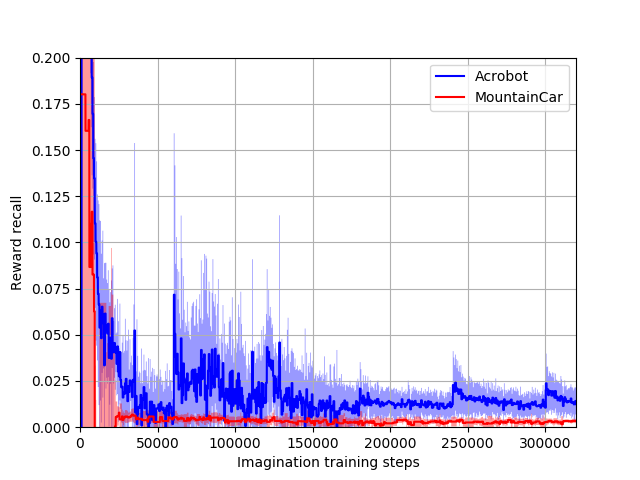}
    \caption{Wasserstein loss for both Acrobot and MountainCar (WGANGP only).} \label{fig:wass}
\end{subfigure} \\%
\begin{subfigure}[b]{.5\linewidth}
    \centering
    \includegraphics[width=1 \textwidth]{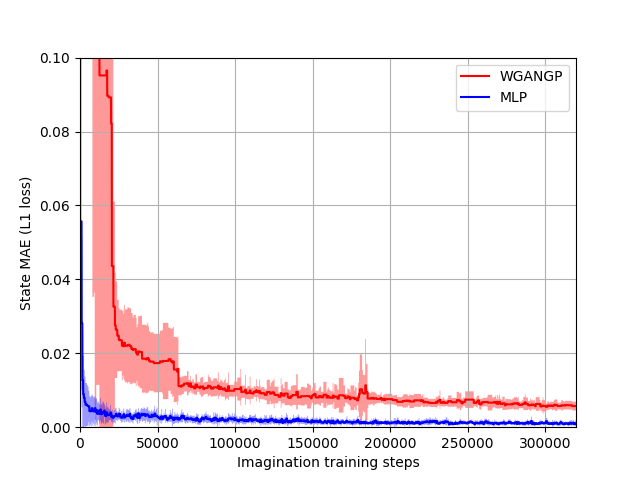}
    \caption{State generation MAE for MountainCar.} \label{fig:l1mountaincar}
\end{subfigure}%
\begin{subfigure}[b]{.5\linewidth}
    \centering
    \includegraphics[width=1 \textwidth]{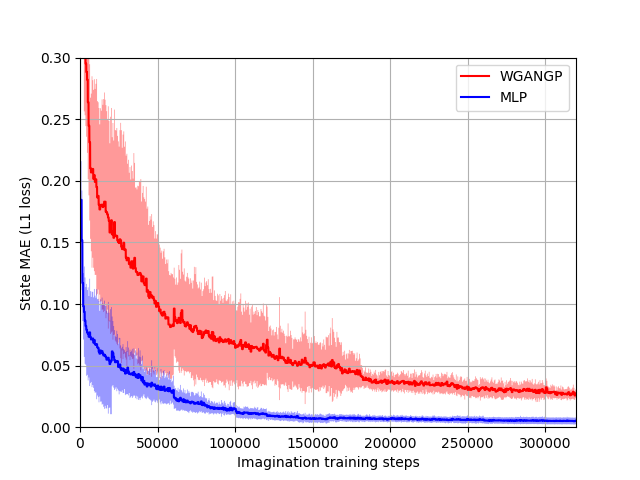}
    \caption{State generation MAE for Acrobot.} \label{fig:l1acrobot}
\end{subfigure}
\caption{Performance of the state-generating imagination submodule on experimental environments. The x-axis represents training steps performed solely in the ITP (in other phases the performance stays constant). The y-axis shows the value of an appropriate metric.}
\label{fig:state}
\end{figure}

The effectiveness of the state imagination submodule is presented in \figurename{ \ref{fig:state}}. Both \figurename{ \ref{fig:l1mountaincar}} and \ref{fig:l1acrobot} show MAE for both WGANGP and MLP state generation. \figurename{ \ref{fig:wass}}, on the other hand, shows Wasserstein distance as approximated by the critic for WGANGP imagination only.

We can see that WGANGP performs worse than the MLP in terms of MAE. This is, however, expected behaviour. MLP optimises for the L1 loss, that is in essence MAE multiplied by a constant, directly. GANs superiority lays in the fact that they do not optimise towards minimising mean error on the individual level, but towards minimising the difference between data distributions. These plots only showcase that indeed both of the modules seem to model dynamics of the real distribution accurately. Using them for direct comparison of the models would be unfair for the WGANGP.

MLP reaches over $99.9\%$ accuracy on MountainCar and $99.5\%$ accuracy on Acrobot, even without much of hypertuning. Unlike with the reward, very high accuracy of state generation is crucial because the correctness of future states highly depends on the correctness of previous states. When making rollouts into the future using pure imagination, errors may compound. Given accuracy $a$ of the state generation model and rollout of length $\tau$, final state's accuracy may drop, in the worst case, to $a^{\tau}$ (nevertheless, this is also unlikely). 

Even WGANGP that does not optimise for MAE reaches good results of over $97\%$ accuracy for both environments proving its convergence properties. This is also shown by the estimated Wasserstein distance between the generator distribution and the environment distribution that reaches values lower than $0.015$ for both environments (even below $0.005$ for MountainCar).

These results show a high promise of replacing the real environment with the imagination, especially considering the fact that not much time and resources have been spent on optimising the architecture and parameters used in the experiments.

\subsection{Data efficiency}
Data efficiency is the main problem targeted in this study. Results in section \ref{sect:imagresults} show that deep learning models can efficiently learn the dynamics of the real environment. The remaining question is whether these models are accurate enough to replace the real environment in the RL process.

\begin{figure}[htb]
\centering
\begin{subfigure}[b]{.5\linewidth}
    \centering
    \includegraphics[width=1 \textwidth]{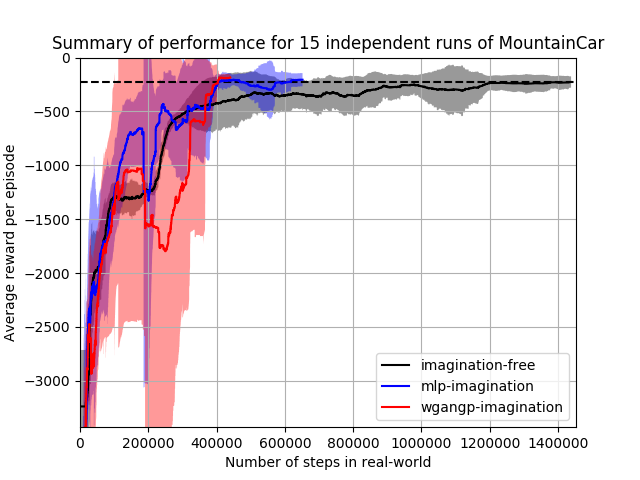}
    \caption{Results on MountainCar}
\end{subfigure}%
\begin{subfigure}[b]{.5\linewidth}
    \centering
    \includegraphics[width=1 \textwidth]{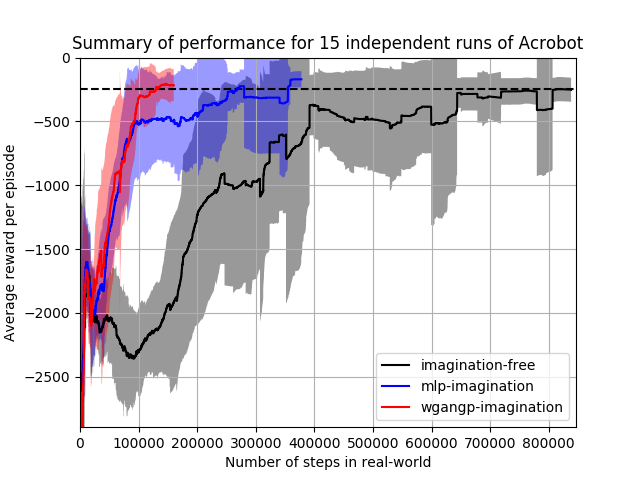}
    \caption{Results on Acrobot}
\end{subfigure}
\caption{Performance of the MLP-based and WGANGP-based framework compared to the state-of-the-art baseline. The x-axis represents a size of experience from the real environment. The y-axis represents a mean reward from past 100 episodes. Black dashed line represents the top performance achieved by the state-of-the-art.} \label{fig:gairlresults}
\end{figure}

\figurename{ \ref{fig:gairlresults}} presents results of both MLP-based and WGANGP-based imaginative framework in comparison to the imagination-free state-of-the-art algorithm. Both imaginative algorithms highly outperform imagination-free Rainbow DQN being more than twice as data efficient on both environments. 

GAIRL requires even as much as over three times fewer experience samples than the framework with MLP-based imagination, and over 6 times less than the current state-of-the-art on the more complex Acrobot environment. Surprisingly, despite the simplicity of MountainCar setting, GAIRL also solves it most optimally.

What is also important, one of the concerns was that imagination-aided agent would be much less stable than the standard model-free algorithm. It is indeed the case on the simple MountainCar, especially for WGANGP-based framework. Remarkably, once the complexity of the environment increases (Acrobot), it is no longer the case.

In addition, two-tailed Wilcoxon signed rank tests were performed to statistically compare obtained results. There were $N=15$ non-zero difference samples in every comparison. The critical value to make sure that the results are statistically significant  ($p \leq 0.05$) for $N=15$ samples is $\omega_c = 25$

Firstly, the comparison between the imagination-free state of the art, and the mlp-based imaginative algorithm was made. For the MountainCar environment, Wilcoxon test produced rank sums $T^+_{m_1} = 114$ and $T^-_{m_1} = 6$. Clearly, $\omega_{m_1} = T^-_{m_1} < \omega_c$. Similarly, in the Acrobot environment rank sums $T^+_{a_1} = 111$ and $T^-_{a_1} = 9$ were collected. Again, $\omega_{a_1} = T^-_{a_1} < \omega_c$. Therefore, the null hypothesis stating that the imaginative framework does not improve sample efficiency of the state of the art can be safely rejected.

Then, the focus moved to the analysis of GAIRL benefits over the more primitive MLP-based imaginative framework. Rank sums for the MountainCar environment in this scenario are $T^+_{m_2} = 82$ and $T^-_{m_2} = 38$. This time, $\omega_{m_2} > \omega_c$. Nevertheless, for the Acrobot, $T^+_{a_2} = 102$ and $T^-_{a_2} = 18$, so $\omega_{a_2} < \omega_c$. Thus, the superiority of GANs in the generative framework is statistically significant only for the more complicated environment. This result was somewhat expected; the central promise of GANs is to work much better on complex and high dimensional domains. The MLP should not be much worse in an elementary environment like MountainCar. 

Full table with the data that was used to calculate test statistics is available in \appendixname{ \ref{app:results}}.

%% file: 2_body/discussion.tex
Two research questions have been posed at the beginning of this thesis: 
\begin{itemize}
    \item Can learning the imaginative model of the environment be more efficient than learning an optimal policy? 
    \item If so, can the learned imagination fulfil the promise of sample efficient model-based RL in settings where dynamics of the real environment are unknown? 
\end{itemize}

Both of the answers have been subsequently answered: Results presented in section \ref{sect:results} clearly show that simply exploiting Markov property allows imagination to converge with almost order of magnitude less data than it is required for learning an optimal policy. Additionally, the imagination was later successfully used to improve on data efficiency of the state-of-the-art Rainbow DQN algorithm.

Initially, the thesis also hypothesised that the GAIRL framework presented throughout the thesis could produce a positive answer to both of the questions. Although in the end it did, one part of the hypothesis has been shown to be redundant in terms of simply answering these questions. Namely, generative adversarial architecture is not a necessity. More traditional models like multilayer perceptron can be successfully deployed within the imaginative framework as well. At least in simple environments used for the evaluation. Nevertheless, although not necessary, GANs tend to produce better results and should scale better to more complex settings.

Two main novel contributions of this thesis would be:
\begin{itemize}
    \item Efficient learning of the environment dynamics (creating imagination) by leveraging Markov property and advantages of generative adversarial networks.
    \item Successful use of imagination to highly improve state-of-the-art data efficiency of deep reinforcement learning through Dyna-Q-inspired algorithm.
\end{itemize}

Unfortunately, a month before the final submission deadline, \citet{kaiser2019model} introduced SimPLe -- a novel deep RL algorithm that follows similar principles to the GAIRL framework. It was also inspired by the Dyna-Q algorithm and produced substantially more data efficient results than the Rainbow DQN. Therefore, it strips the novelty out of the second contribution of this thesis. However, because SimPLe leverages traditional L1/L2 loss model proposed by \citet{oh2015ataripred} with only a few minor modifications, it often suffers from inaccurate generation (blurry images, disappearing small but crucial features). GAIRL's approach of using GANs promises to circumvent this difficulty. However, more benchmarks in more complex environments should be performed to prove this hypothesis fully. This brings us to the point in the next paragraph.

GAIRL seems to overperform most recent state-of-the-art and even introduce crucial contribution on top of the similar advancement that was produced by the top research institution in parallel. Nevertheless, the set of environments used for testing was limited due to strict time constraints. The framework needs to be evaluated on a higher variety of domains. Each of them should reflect one or more of the following properties: a very high dimensional state space, non-deterministic transition dynamics, or a partially observable Markov decision process (POMDP). GANs architecture promises to handle the first two smoothly, theoretically amplifying the gap between GAIRL and other approaches. However, GAIRL may perform worse in environments following the POMDP properties. Additionally, it should be empirically compared with the recently introduced SimPLe algorithm. Further experiments are the critical next step and are planned to continue after the submission of this thesis.

What is more, more detailed optimisation of GAIRL architecture has been left for future work. Not much focus has been given to hyperparameters. It would be interesting to see a thorough study of the best parameter choices for the GAIRL setting. Potentially, sample efficiency could have been even more significantly improved by decreasing the length of the model-free phase and compensating it with the agent's training in the imagination-based phase. Additionally, reward and state generation should be combined within a single machine learning model. It is especially essential for the reward generation so it can utilise the benefits of GANs in stochastic environments. 

Another promising direction is to modify GAIRL's memory to follow a similar structure to the prioritised replay buffer \citep{schaul2015prioritized} that is used in the Rainbow DQN. It could lead to intriguing results. Moreover, current use of Wasserstein GAN could be utilised to guide the exploration-exploitation trade-off. As \citet{azizzadenesheli2018ganmcts} mentioned, high Wasserstein distance produced by the critic for certain states could indicate that the agent is unsure of the possible outcomes of such a state. Therefore, the agent should potentially move there to explore the search space better, even if it seems less optimal.

Finally, the length of the imagination training phase $\rho_2$ could become adaptive. Currently, it takes a fixed constant. However, imagination does not need the same amount of training time at every iteration. The first iteration should take the longest, as the imagination starts without any prior knowledge. However, during the second iteration, it already has a general overview of the world. It just needs to update its model slightly to capture newly gathered data-points. Consequently, by the law of large numbers, it could follow that if $n$ indicates the number of iterations then $\lim_{n \to \inf} \rho_2 = 0$, potentially saving a substantial amount of computational power.

One of the main issues raised during the project demonstration session was that the presented comparison of GAIRL with Rainbow DQN is not fair towards the latter. The argument was that the GAIRL framework performs a considerable amount of additional computation in the background. The fact that GAIRL requires more computation is true. Although it has the same time complexity in terms of the big O notation, it performs slower and is less computationally efficient as shown in \appendixname{ \ref{app:compeff}}. Naturally, using the real environment directly instead of its imperfect imagination allows the agent to find the optimal policy quicker, even without mentioning the computational power required to learn the imagination.

Nonetheless, data efficiency, not computational efficiency, is one of the most pressing problems in the field \citep{irpan2018rlfails}, and the focus of this thesis. Computational power is cheap and widely accessible. Performing millions of random trial and error actions in the real environment, however, is often either very expensive or even impossible as explained throughout the thesis. Furthermore, this type of comparison is not new to the field. It has been used repeatedly in the literature \citep{mnih2015dqn, silver2016alphago, silver2017alphago, hessel2017rainbow, schulman2017ppo, kaiser2019model}

Another argument was that the imaginative module would not be able to grasp the dynamics of complex environments. Example of the real world in case of robotics was given. It is a reasonable concern. We cannot know for sure before performing appropriate experiments in such an environment as already mentioned earlier. However, the imagination module requires the same assumptions and the same setting as the current state-of-the-art RL algorithms. In theory, if imagination cannot encapsulate the real environment, then the state-of-the-art is not able to learn a close-to-optimal policy either. Notwithstanding, this question is open for future experimental work.

The last point referred to employed benchmark environments. Namely, what is the point of learning the imaginative simulation for the environment that already is a simulation of a car/robot? Undeniably, it does not make sense in practice. However, the goal of this research was not to find the best solution to the Acrobot or MountainCar. It was about improving the general-use state-of-the-art reinforcement learning. Simulations were used merely as a mean to compare the capabilities of presented RL algorithms. Optimal or close-to-optimal solutions to these environments, much better than any RL algorithm, has been devised decades ago. It did not stop, however, RL researchers to use them for benchmark purposes on conferences such as NeurIPS or ICML.

%% file: 2_body/conclusion.tex
The goal of this thesis was to improve data efficiency of the state-of-the-art reinforcement learning. This has been successfully achieved by introducing an imaginative framework that can accurately and efficiently approximate dynamics of the real environment by making use of Markov property and generative adversarial models.

It presented experimental evidence that supports the superiority of GANs over standard generative models for conditional state prediction within the reinforcement learning setting. It also introduced a way to utilise imperfect approximation of the real world to limit the amount of data needed to train an optimally behaving agent.

Similar advancement regarding sample-efficient reinforcement learning has been released in parallel \citep{kaiser2019model}. The study presented here, however, not only confirms the results obtained by \citet{kaiser2019model} but also adds an important contribution to the field that promises to improve the state-of-the-art even further.

Nevertheless, more experiments are needed to ensure the superiority of the introduced framework, and there is still room open for future work.

%% file: 3_appendices/results.tex
\begin{table}[H]
\centering
\hspace*{-0.1\linewidth}
\begin{tabular}{|l|l|l|l|l|} \hline
    Environment  &  Random seed& Imagination-free          & MLP imagination        &  WGANGP imagination\\ \hline \hline
    MountainCar  &  10         & 490.1                   & 856.1               & 292.6            \\
    MountainCar  &  50         & 1314.4                  & 625.2               & 255.7            \\
    MountainCar  &  100        & 821.5                   & 397.3               & 300.3            \\
    MountainCar  &  500        & 514.6                   & 297.2               & 325            \\
    MountainCar  &  1000       & 970.2                   & 415.3               & 395.8            \\
    MountainCar  &  5000       & 510.8                   & 245.6               & 134            \\
    MountainCar  &  10000      & 950.5                   & 222.3               & 317.8            \\
    MountainCar  &  50000      & 516.3                   & 372.6               & 336.1            \\
    MountainCar  &  100000     & 1156                    & 390.9               & 556.5            \\
    MountainCar  &  500000     & 881.3                   & 531.5               & 236            \\
    MountainCar  &  1000000    & 726.1                   & 298.3               & 217.1            \\
    MountainCar  &  5000000    & 687.9                   & 207.7               & 210.6           \\
    MountainCar  &  10000000   & 791.7                   & 456.1               & 411            \\
    MountainCar  &  50000000   & 1095                    & 369                 & 335.5            \\
    MountainCar  &  100000000  & 935.2                   & 377.2               & 533.3            \\
    \hline
    Acrobot      &  10         & 641.2                   & 205.1              & 71.2            \\
    Acrobot      &  50         & 814.5                   & 74.3               & 94.4              \\
    Acrobot      &  100        & 328.4                   & 118.1              & 71.4            \\
    Acrobot      &  500        & 153.6                   & 152.7              & 163.3             \\
    Acrobot      &  1000       & 548.8                   & 75.8               & 150            \\
    Acrobot      &  5000       & 207.5                   & 96.3               & 119.9            \\
    Acrobot      &  10000      & 254.7                   & 87                 & 65.1            \\
    Acrobot      &  50000      & 601                     & 317.6              & 132.1            \\
    Acrobot      &  100000     & 176.2                   & 359.7              & 100            \\
    Acrobot      &  500000     & 653.2                   & 230.2              & 151.6            \\
    Acrobot      &  1000000    & 285.9                   & 291.1              & 124.3            \\
    Acrobot      &  5000000    & 399.7                   & 117.9              & 101.3            \\
    Acrobot      &  10000000   & 397.2                   & 199.3              & 74.1            \\
    Acrobot      &  50000000   & 318.1                   & 172                & 86.8            \\
    Acrobot      &  100000000  & 324.1                   & 139.7              & 68.8            \\\hline
    \hline
\end{tabular}
\captionsetup{width=1.15\linewidth}
\caption{Algorithms comparison. Runs are grouped by the environment and used random seed. Values in the columns represent the number of samples from the real environment needed before convergence.} 
\label{table:time}
\end{table}

%% file: 3_appendices/computations.tex
\tablename{ \ref{table:time}} presents a comparison of the algorithms in regards to the time they required until convergence on a 4CPU instance. For both Acrobot and MountainCar, imaginative framework employing MLP for the imagination module requires approximately 7 times more time to converge to the optimal policy than the imagination-free agent. Full GAIRL algorithms run approximately 2.5 times longer than the MLP-based agent and 17 times longer than the imagination-free agent.

It clearly shows that the state-of-the-art is much more computationally efficient than any of the proposed algorithms. Combining it with the results from section \ref{sect:results}, we can observe that there is an inverse proportional relation between data efficiency and computational efficiency.

However, GAIRL performance should scale better with higher computational power. The most demanding part of GAIRL is the imagination training phase. The ITP employs a standard supervised learning process that can easily leverage high parallelism provided by GPUs and TPUs. Highly sequential reinforcement learning, on the other hand, is much harder to parallelise. Naturally, GAIRL can never reach the same efficiency as imagination-free options; however, it may get asymptotically closer when more powerful hardware is provided.

\begin{table}[H]
\centering
\hspace*{-0.1\linewidth}
\begin{tabular}{|l|l|l|l|l|} \hline
    Environment  &  Random seed&   WGANGP imagination      & MLP imagination       & Imagination-free  \\ \hline \hline
    MountainCar  &  10         & 464.534                   & 298.221               & 49.078            \\
    MountainCar  &  50         & 822.961                   & 304.695               & 55.173            \\
    MountainCar  &  100        & 651.695                   & 212.879               & 29.919            \\
    MountainCar  &  500        & 1047.146                  & 414.624               & 31.019            \\
    MountainCar  &  1000       & 334.610                   & 232.937               & 46.608            \\
    MountainCar  &  5000       & 652.493                   & 340.321               & 33.857            \\
    MountainCar  &  10000      & 333.088                   & 246.786               & 35.686            \\
    MountainCar  &  50000      & 285.082                   & 133.015               & 42.096            \\
    MountainCar  &  100000     & 906.235                   & 375.812               & 32.433            \\
    MountainCar  &  500000     & 761.129                   & 111.714               & 59.211            \\
    MountainCar  &  1000000    & 615.542                   & 286.956               & 31.195            \\
    MountainCar  &  5000000    & 741.330                   & 381.619               & 42.303            \\
    MountainCar  &  10000000   & 936.039                   & 152.674               & 41.251            \\
    MountainCar  &  50000000   & 695.203                   & 166.715               & 36.890            \\
    MountainCar  &  100000000  & 603.739                   & 225.416               & 35.841            \\
    \rowcolor[HTML]{C0C0C0} 
    Mean         &  --         & 656.722                   & 258.959               & 40.171            \\ \hline \hline
    
    Acrobot      &  10         & 119.521                   & 138.605               & 13.962            \\
    Acrobot      &  50         & 104.571                   & 84.227                & 41.711            \\
    Acrobot      &  100        & 312.638                   & 216.284               & 20.661            \\
    Acrobot      &  500        & 536.062                   & 192.328               & 8.235             \\
    Acrobot      &  1000       & 334.610                   & 127.010               & 26.104            \\
    Acrobot      &  5000       & 485.868                   & 96.694                & 10.359            \\
    Acrobot      &  10000      & 333.088                   & 209.477               & 22.162            \\
    Acrobot      &  50000      & 152.203                   & 70.388                & 10.528            \\
    Acrobot      &  100000     & 259.356                   & 185.065               & 30.179            \\
    Acrobot      &  500000     & 605.006                   & 28.556                & 19.241            \\
    Acrobot      &  1000000    & 262.557                   & 125.856               & 12.956            \\
    Acrobot      &  5000000    & 116.214                   & 171.756               & 11.067            \\
    Acrobot      &  10000000   & 335.291                   & 143.612               & 14.851            \\
    Acrobot      &  50000000   & 224.329                   & 91.233                & 26.151            \\
    Acrobot      &  100000000  & 313.412                   & 185.561               & 19.451            \\
    \rowcolor[HTML]{C0C0C0} 
    Mean         &  --         & 335.155                   & 137.777               & 19.175            \\ \hline
    
    \hline
\end{tabular}
\captionsetup{width=1.15\linewidth}
\caption{Algorithms comparison. Runs are grouped by the environment and used random seed. Values in the columns represent the number of minutes that passed before convergence. Experiments run on Google Cloud Platform, each on the instance with 4 Intel Skylake CPUs and 8GB of RAM.} 
\label{table:time}
\end{table}

%% file: 3_appendices/hyperparameters.tex
\begin{table}[H]
\centering
\begin{tabular}{|l|l|}
    \hline
    Hyperparameter                               & Value              \\ \hline
    Hidden layers                                & [256, 512, 1024]   \\
    Hidden layers activation                     & Leaky ReLU         \\
    Leakiness parameter ($\alpha_{lrelu}$)       & $0.2$              \\
    Dropout probability                          & $0$                \\
    Final layer activation                       & Tanh               \\
    Optimiser                                    & Adam               \\
    Learning rate ($\alpha_{lr}$)                & $2 \times 10^{-4}$ \\
    First Adam decay rate ($\beta_1$)            & $0.9$              \\
    Second Adam decay rate ($\beta_2$)           & $0.999$            \\ \hline
\end{tabular} 
\caption{Final parameters of the multilyer perceptron used for learning the generative distribution of the MNIST dataset.} 
\end{table}

\begin{table}[H]
\centering
\begin{tabular}{|l|l|}
    \hline
    Hyperparameter                               & Value              \\ \hline
    Generator layers                             & [256, 512, 1024]   \\
    Generator final layer activation             & Tanh               \\
    Critic layers                                & [1024, 1024, 1024] \\
    Critic final layer activation                & Linear             \\
    Critic steps for one generator step          & $10$               \\
    Hidden layers activation                     & Leaky ReLU         \\
    Leakiness parameter ($\alpha_{lrelu}$)       & $0.2$              \\
    Dropout probability                          & $0$                \\
    Noise size                                   & $100$              \\
    Penalty coefficient                          & $10$               \\
    Optimiser                                    & Adam               \\
    Learning rate ($\alpha_{lr}$)                & $2 \times 10^{-4}$ \\
    First Adam decay rate ($\beta_1$)            & $0.5$              \\
    Second Adam decay rate ($\beta_2$)           & $0.9$              \\ \hline
\end{tabular} 
\caption{Final parameters of the Wasserstein GAN with Gradient Penalty used for learning the generative distribution of the MNIST dataset.} 
\end{table}

\begin{table}[H]
\centering
\begin{tabular}{|l|l|}
    \hline
    Hyperparameter                               & Value              \\ \hline
    Generator layers                             & [256, 512, 1024]   \\
    Generator dropout probability                & $0$                \\
    Generator final layer activation             & Tanh               \\
    Discriminator layers                         & [1024, 512, 256]   \\
    Discriminator dropout probability            & $0.2$              \\
    Discriminator final layer activation         & Sigmoid            \\
    Discriminator steps for one generator step   & $1$                \\
    Hidden layers activation                     & Leaky ReLU         \\
    Leakiness parameter ($\alpha_{lrelu}$)       & $0.2$              \\
    Noise size                                   & 100                \\
    Optimiser                                    & Adam               \\
    Learning rate ($\alpha_{lr}$)                & $2 \times 10^{-4}$ \\
    First Adam decay rate ($\beta_1$)            & $0.9$              \\
    Second Adam decay rate ($\beta_2$)           & $0.999$            \\ \hline
\end{tabular} 
\caption{Final parameters of the original GAN used for learning the generative distribution of the MNIST dataset.} 
\end{table}